\documentclass{article}

 \usepackage[preprint]{neurips_2025}

\usepackage[utf8]{inputenc} %
\usepackage[T1]{fontenc}    %
\usepackage{url}            %
\usepackage{booktabs}       %
\usepackage{amsfonts}       %
\usepackage{nicefrac}       %
\usepackage{microtype}      %
\usepackage{xcolor}         %
\usepackage{comment}
\usepackage{booktabs}
\usepackage{pifont}    
\usepackage{colortbl}
\usepackage[inline]{enumitem}
\usepackage{multirow}
\usepackage{comment}
\usepackage{subcaption}
\newcommand{\cmark}{\textcolor{green!70!black}{\ding{51}}}
\newcommand{\xmark}{\textcolor{red!70!black}{\ding{55}}}

\newcommand{\cellg}{\cellcolor{green!10}}
\newcommand{\cellr}{\cellcolor{red!10}}

\usepackage{amsmath}
\usepackage{amssymb}
\usepackage{graphicx}
\usepackage[colorlinks=true, allcolors=blue]{hyperref}
\usepackage[capitalize,noabbrev]{cleveref}
\usepackage{physics}
\usepackage{bm}
\usepackage{comment}
\usepackage{tikz}
\usepackage{tikz-dimline}
\usetikzlibrary{plotmarks,calc,arrows.meta,patterns,shapes.arrows,external,decorations,shapes,positioning,fit,backgrounds}
\tikzset{>=latex}

\usepgfplotslibrary{groupplots,dateplot}
\usetikzlibrary{angles,quotes}
\usepgfplotslibrary{colormaps}
\usetikzlibrary{pgfplots.colormaps}
\usepgfplotslibrary{fillbetween}
\pgfplotsset{compat=newest}

\newcommand\p{\ensuremath{\partial}}

\title{A fully GPU-based workflow for building physics emulators of hypersonic flows}

\author{
  \textbf{Fabian Paischer}\thanks{Equal contribution}$\:\:^{2,3}$\quad
  \textbf{Dylan Rubini}\footnotemark[1]$\:\:^{3}$\quad
  \textbf{Deniz A. Bezgin}$^{1}$\quad
  \textbf{Aaron B. Buhendwa}$^{1}$\quad\\
  \textbf{David Hauser}$^{3}$ \quad
  \textbf{Florian Sestak}$^{2,3}$\quad
  \textbf{Johannes Brandstetter}$^{2,3}$\quad
  \textbf{Sebastian Kaltenbach}$^{3}$\\
  \textbf{Nikolaus A. Adams}$^{1}$
  \\[8pt]
  {$^1$~Chair of Aerodynamics and Fluid Mechanics, TU Munich, Germany}\\
  {$^2$~ELLIS Unit, Institute for Machine Learning, JKU Linz}\\
  {$^3$~EMMI AI, Linz}\\
}

\begin{document}

\maketitle

\begin{abstract}

The ability to resolve complex physical phenomena with high fidelity and at low computational cost is central to addressing key challenges in modern engineering. 
A prime example lies in hypersonic flows, where the precise prediction of the full flowfield topology, in particular with respect to shock wave location and intensity, is critical. 
Yet supersonic and hypersonic flows continue to be a stumbling block for traditional reduced-order models and neural emulators that struggle to capture steep gradients in flow states with physical consistency in applications of industrial relevance. 
To that end, we introduce a fully GPU based workflow that integrates accelerated data generation with the training of neural emulators augmented by uncertainty quantification and physics-aware refinement.
Our workflow is enabled by a differentiable high-fidelity solver (JAX-Fluids) which we employ for rapid dataset creation and residual-based improvement of the neural emulator to enhance physical consistency. 
Building on this framework, we first present a suite of model architectures and analyze their scaling behavior to expose their strengths and shortcomings. 
We then show that residual-based refinement enables training on cases where only mesh and input parameters are available, substantially reducing residuals and improving physical consistency.
Together, differentiable simulation and residual-based refinement yield physics emulators that remain reliable beyond their training distribution, a key requirement for deploying surrogates in real-world engineering design loops.

\end{abstract}

\section{Introduction}

Fluid dynamics is fundamental to many processes in nature and technology, and its numerical simulation routinely ranks among the top-level compute-resource allocations at Tier~0 computing facilities \cite{slotnick2014cfd}.
Despite its complex and inherently multi-scale character, flow fields exhibit coherent behavior characterized by universal dependencies.
A long-standing challenge for predictive simulation, particularly in high-speed transport and propulsion, is the simultaneous presence of multi-scale flow structures such as coherent vortices, eddies, and shocks.
Shockwaves are among the most consequential of these phenomena, governing processes from the evolution of galaxies \cite{McKee80} to the feasibility of high-speed propulsion \cite{Urzay18}.
They are characterized by discontinuities in the macroscopic flow state, such as extreme gradients in pressure, density, temperature, and momentum, which arise when the local flow speed $\|\mathbf{u}(\mathbf{x},t)\|$ exceeds the local speed of sound $a(\mathbf{x},t)$, i.e., when the local Mach number $Ma(\mathbf{x},t) = \|\mathbf{u}\| / a \geq 1$.

Hypersonic flowfield predictions are particularly challenging for both numerical simulation and data-driven modeling.
They are characterized by Mach numbers beyond approximately five and exhibit strong shock interactions, high-enthalpy effects, and stringent conservation requirements.
Such flow problems have historically served as driving applications for high-performance computing, motivating both methodological advances and community benchmarks \cite{Wilfong_ETAL_25,Rossinelli_ETAL_13}.
The computational cost of numerical simulations has made data-driven alternatives to classical computational fluid dynamics (CFD) a foremost research interest \cite{BruntonNoackKoumoutsakos20, brenner2019perspective, karniadakis2021physics}. 
These range from full surrogate substitution of the numerical solver via Physics Emulators (PEs) to machine-learned (ML) acceleration of existing CFD methodologies.

In this work, we present a fully GPU-based workflow for hypersonic flows enabled by the differentiable finite-volume solver JAX-Fluids \citep{BezginBuhendwaAdams23,BezginBuhendwaAdams25}.
The proposed workflow encompasses
\begin{enumerate*}[label=(\roman*)]
    \item GPU-accelerated data generation,
    \item pre-training of neural emulators, and
    \item target-free, residual-based refinement.
\end{enumerate*}
Data generation is based on a Cartesian multi-block mesh that enables efficient parallelization on GPU and seamless integration with various ML architectures and training paradigms.
For pre-training, we investigate two complementary architectures, namely irregular-grid based AB-UPT \citep{alkin2025ab} and regular-grid based vision transformer \citep[ViT,][]{dosovitsky2021vit}.
Furthermore, we investigate the trade-off between deterministic and probabilistic training paradigms.
We conduct scaling studies with respect to both model size and dataset size for all architectures and training paradigms.
For target-free residual-based refinement the pre-trained PE generates a candidate solution which is evaluated against the residual of the underlying partial differential equation (PDE) computed by the differentiable solver.
The resulting gradient signal is backpropagated into the emulator weights without requiring any target flow fields, mirroring exactly the numerical discretization used during data generation.

Our experiments reveal several key findings for architecture and training paradigm selection and refinement for PEs in the hypersonic regime.
Among the two architectures, AB-UPT achieves the highest accuracy in data-abundant settings, while the ViT outperforms in data-scarce regimes due to the strong inductive bias provided by its regular-grid structure.
Flow matching trades point-wise accuracy for generative modeling capability, but  provides off-the-shelf uncertainty and acts as implicit data augmentation, yielding a smaller gap between in-distribution and out-of-distribution performance than either deterministic architecture.
For physics-aware refinement, we find that backpropagation through the PDE residual leads to substantial reductions in conservation residuals with little to no changes in field-level accuracy, suggesting that the pre-trained models already capture the dominant flow structure and the refinement primarily corrects local physical consistency.
Notably, the target-free setup, which conditions on the computational mesh and input parameters without requiring reference flow fields exhibits the largest improvement in residuals.

Overall our contributions are as follows.\begin{itemize}[leftmargin=*]
    \item We present a %
    fully GPU-based workflow for hypersonic flow emulation that integrates data generation, surrogate pre-training, and physics-aware refinement within a single differentiable pipeline built on the JAX-Fluids solver.
    \item We enable regular-grid ML architectures to be agnostic to grid topology parameters (e.g., block count and ordering of the block-structured meshes) by combining absolute and relative positional encodings based on coordinates in physical space. 
    \item We evaluate two complementary neural architectures (AB-UPT and ViT) and two training paradigms (deterministic vs probabilistic) and conduct scaling studies with respect to both model size and dataset size, identifying distinct data-efficiency and accuracy trade-offs across regimes.
    \item We introduce a target-free refinement stage that improves physical consistency by backpropagating PDE residuals into the pre-trained neural PE weights without requiring reference flow fields, and demonstrate its advantage over field-value fine-tuning.
\end{itemize}

\section{Related Work}
\label{sec:related_work}

Machine learning has been integrated into CFD workflows in several complementary ways. 
One line of work augments classical numerical schemes with learned components.
Neural networks (NNs) have served as troubled-cell indicators that locate where limiting is needed in high-order discretizations \cite{ray2018artificial}, as local and parsimonious modifications within physics-constrained implicit Large-Eddy Simulation (LES) that lead to modifications of classical shock-capturing schemes \cite{Bezgin_ETAL_25}, and as learned correction operators that recover fine-grid accuracy from coarse-grid solvers
\cite{kochkov2021machine}. 
A related line uses reinforcement learning to discover effective closures \cite{novati2021automating,fischer2025optimal}. 
A third direction bypasses the solver entirely and trains neural networks as end-to-end PEs, e.g., deep convolutional models that map airfoil geometry directly to Reynolds-averaged Navier--Stokes (RANS) fields \cite{thuerey2020deep}. 
A common enabler across all of these settings is the availability of a state-of-the-art CFD solver with algorithmic differentiation capability for end-to-end pipelines \cite{BezginBuhendwaAdams23,BezginBuhendwaAdams25}.

A more general formulation seeks to learn the parameter-to-solution operator of a PDE as a mapping between function spaces. 
Neural operators provide a scalable and resolution-invariant framework for learning mappings and offer orders-of-magnitude speedups over traditional numerical solvers \cite{li2020fourier,lu2021learning,Azizzadenesheli_ETAL_24}. 
Additionally, for unstructured discretizations, mesh-based graph networks exploit the adjacency structure of the simulation grid to learn local update rules \cite{pfaff2020learning}. 
Recently, purely transformer-based \citep{vaswani2017attention} formulations \cite{alkin2024universal,alkin2025ab} have shown to successfully scale to industry relevant complexity and effectively capture long range dependencies. 
Despite their flexibility, neural PEs still face well-documented practical limitations, for example heavy data requirements, sensitivity to training-distribution shifts, lack of rollout robustness, and the absence of guaranteed physical consistency in complex regimes \cite{VinuesaBrunton22}.
Most of the current work therefore has focused on optimizing model errors, for instance RANS equations \cite{GuptaDuraisamy26}.
Questions of scalability and applicability beyond such regimes, including hypersonic flight, remain open.

Although small-scale fluctuations in velocity, pressure, and density can in principle be resolved on sufficiently fine grids, shocks remain genuine discontinuities and require dedicated nonlinear schemes \cite{Toro09}. 
Shocks are weak solutions of the underlying flow equations and obey precise jump relations between pre- and post-shock states, namely the Rankine--Hugoniot conditions \cite{LeVeque92}. 
These properties make reliable, high-resolution prediction of shocks and shock interactions challenging even for classical schemes and notoriously difficult for purely data-driven surrogates. 
Only with the explicit inclusion of inductive biases for physical consistency have physics-informed neural networks \citep{karniadakis2021physics} delivered predictions with correct shock locations, satisfied jump conditions, and maintained positivity of the flow state \cite{mao2020physics, JagtapMaoAdamsKarniadakis22}. 
The optimization of a discrete loss provided by a numerical discretization of the governing equations on a chosen mesh has more recently proven sufficiently effective to address inverse inference problems in three-dimensional steady-state transonic and supersonic flows \cite{Buhendwa_ETAL_25,paischer2025going}, although in that setting some flow-field data must still be available.

For generative tasks, denoising diffusion models have begun to be explored in fluid mechanics. 
They produce sample-diverse predictions and naturally provide a posterior from which uncertainty can be estimated \cite{ho2020denoising}. 
Diffusion models enable fast forecasting of distributional quantities of interest in high-dimensional dynamical systems \cite{gao2024generative,molinaro2024generative}. 
In incompressible turbulence, diffusion models have generated
physically plausible three-dimensional flow states from scratch
\cite{lienen2024zero} and delivered calibrated uncertainty for airfoil flows over a range of Reynolds numbers and angles of attack \cite{liu2024uncertainty}. 
In the compressible regime, denoising diffusion models have been examined for moderately supersonic flow \cite{AbaidiAdams25}.
However, a fully GPU-based workflow that combines parameterized data generation, neural emulator training across complementary architectures, uncertainty quantification, and physics-aware refinement for complex hypersonic flows has not yet been demonstrated.

\section{The Neural Physics Emulator Pipeline}
\label{sec:pipeline}

\begin{figure}[h!]
    \centering
    \resizebox{\linewidth}{!}{%
        \input{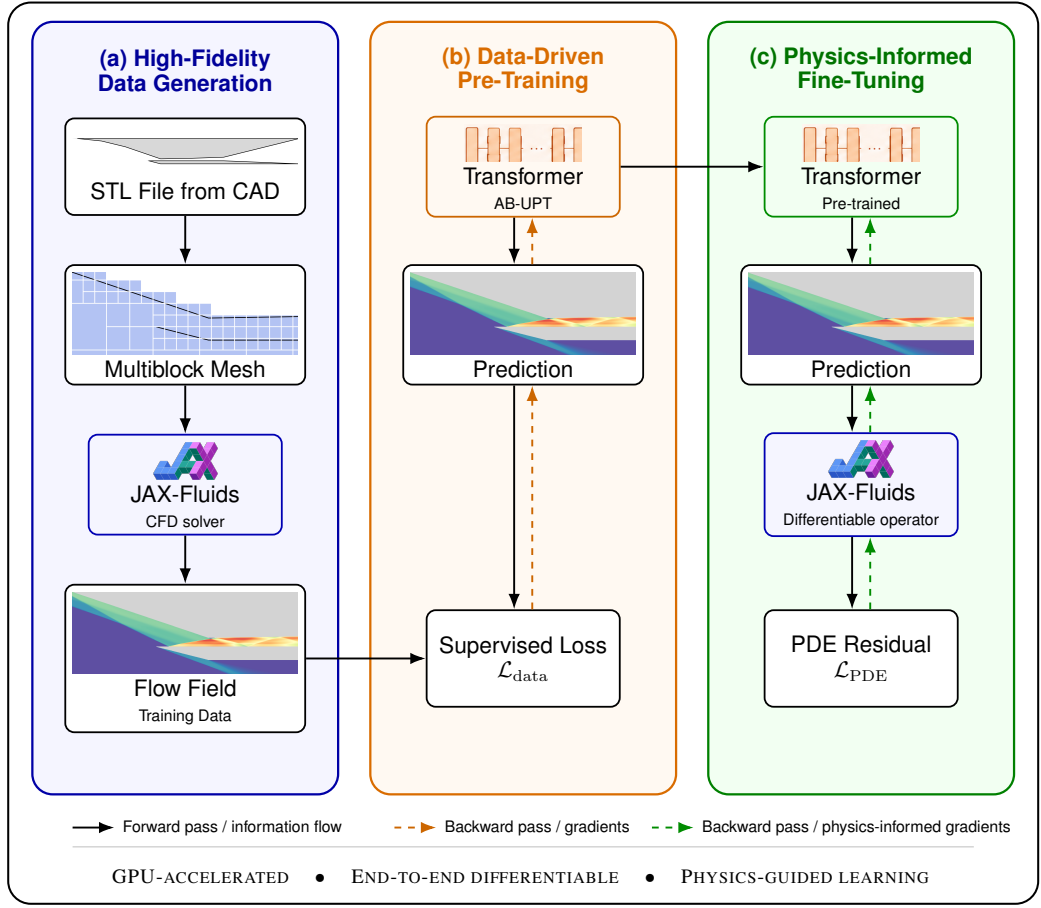}%
    }
    \caption{
    A fully GPU-accelerated and end-to-end differentiable workflow for constructing physics emulators of complex flow phenomena. The workflow consists of three stages. (a) High-Fidelity Data Generation: Starting from an STL representation of the geometry, a multi-block mesh is automatically generated. JAX-Fluids then performs high-fidelity CFD simulations until a steady-state solution is obtained, yielding the training dataset. (b) Data-Driven Pre-Training: Physics emulators are pre-trained in a supervised manner using the high-fidelity dataset generated in stage (a). (c) Physics-Informed Fine-Tuning: After pre-training, the emulators are refined in a target-free manner by minimizing the residuals of the governing equations. The residuals are computed by evaluating the differentiable JAX-Fluids solver on the model predictions, enabling end-to-end gradient-based optimization through the model and the CFD solver itself.   
}
\label{fig:overview}
\end{figure}

Drawing inspiration from the success of Large Language Models (LLMs) in natural language processing \cite{achiam2023gpt,team2023gemini}, we propose a workflow for building PEs that ranges from data generation to pre-training and fine-tuning. 
While LLMs excel at modeling linguistic structures, a PE is specifically designed to learn the complex functional relationships of a physical system. 
Despite their differences, both LLMs and PEs can be based on the same underlying attention mechanism \cite{vaswani2017attention}.

We deliberately choose the term \textit{Physics Emulator} to distinguish this work from generic black-box regression models. 
The term \textit{Physics} signals that the systems of interest are physical and not just arbitrary input-output mappings. 
The term \textit{Emulator} is adopted from the statistical literature on computer experiments, where it denotes a fast, probabilistic surrogate trained on the outputs of a computationally expensive simulator \cite{sacks1989design, kennedy2001bayesian}. 
In that tradition, an emulator is not merely a curve fit, it is a high-fidelity statistical proxy designed to reproduce the full input-output behavior of the underlying code. 
Consequently, a Physics Emulator is such a surrogate purpose-built for physical simulators, functioning as a drop-in replacement for solvers such as CFD, mapping input parameters to complete physical fields in milliseconds rather than hours.
 
To effectively serve as a neural surrogate for hypersonic applications, we identify the following desirable characteristics for a PE, the first two of which we consider
essential:
\begin{itemize}
\item \textbf{Differentiability}: The model must support end-to-end automatic differentiation for seamless integration into gradient-based design optimization and for physics-based fine-tuning using differentiable CFD.
\item \textbf{Physical consistency}: Predictions must respect the conservation laws of mass, momentum, and energy as much as possible.
\item \textbf{Uncertainty capabilities:} When predictive uncertainty is relevant 
for the downstream task, the PE should be able to provide a posterior distribution 
that can be sampled to yield uncertainty estimates, rather than only point 
predictions.
\end{itemize}
To construct a Physics Emulator that satisfies these requirements, we propose a fully GPU-resident pipeline comprising three phases, see \cref{fig:overview}:
 
\begin{enumerate}
\item \textbf{Data Generation:} First, data generation requires defining a robust design space. 
This involves parameterizing the geometries, material properties and boundary conditions of the target engineering system.
By establishing a comprehensive parametric envelope, we ensure the resulting model will be exposed to a diverse and representative set of physical scenarios. 
Based on the parameterized setup, high-fidelity data must be generated, which requires a scalable and accurate solver. 
\item \textbf{Model Training:} Given a dataset of high-fidelity simulations, we pre-train a neural PE. 
This step includes the selection of a suitable model architecture and training process.
For instance, in case predictive uncertainty should be quantified a probabilistic framework is required.
\item \textbf{Model Fine-tuning:} The base model can be fine-tuned based on specific quantities of interest (e.g., conservation laws). 
These quantities can be chosen after pre-training and can be done in a target-free manner.
\end{enumerate}

The following sections discuss all three phases in detail:

\subsection{Data Generation: JAX-Fluids}
\label{subsec:jax-fluids}

Within the scope of this work, high-fidelity flow-field data are generated using JAX-Fluids~\cite{BezginBuhendwaAdams23,BezginBuhendwaAdams25}, a high-order, fully differentiable finite-volume solver for compressible single- and two-phase flows.
JAX-Fluids combines high-order shock-capturing discretizations, GPU-acceleration, automated Cartesian multi-block meshing, and end-to-end automatic differentiation within a single JAX-based framework, making it particularly well suited for training of PEs. In particular, 
these properties enable using the solver not only as an offline data generator, but also as a differentiable physics engine during model training and downstream optimization.

Hypersonic flows are characterized by complex flow phenomena like shock-shock interactions, shock-interface interactions, wave dynamics, viscous-inviscid interactions, flow separation, and multi-species effects.
Accurately resolving these phenomena requires numerical methods that are both robust in the presence of discontinuities and sufficiently accurate in smooth regions of the flow. 
In this work, the data are generated in the inviscid limit governed by the compressible Euler equations which corresponds to typical application scenarios of high Reynolds numbers. 
At hypersonic Mach numbers considered here, the intake flowfield and the integral performance metrics are dominated by the shock structure which the Euler equations admit as weak solutions satisfying Rankine–Hugoniot jump conditions. 
For hypersonic intakes operating at flight Reynolds numbers, viscous regions are confined to relatively thin boundary layers and inviscid analysis is the standard scope for preliminary scramjet-inlet design \cite{heiser1994hypersonic}. 
The PE and refinement methodology described in subsequent section is not specific to the Euler equations and extends to the full Navier–Stokes system by adding viscous effects.

JAX-Fluids follows a high-order Godunov-type finite-volume formulation. 
Shock waves are captured using nonlinear solution-adaptive reconstruction together with approximate Riemann solvers.
For the cases considered in this work, steady-state solutions are obtained by explicit time advancement until the residuals of the governing equations fall below a prescribed tolerance.
The design of JAX-Fluids is motivated by the requirements of ML workflows for computational physics. 
In particular, three aspects are central to the present work: Cartesian multi-block meshes, GPU acceleration, and automatic differentiability.

\paragraph{Cartesian multi-block mesh.}
A central requirement for large-scale training-data generation is a mesh-generation procedure that is robust, automated, and computationally efficient. In general, CFD mesh strategies can be classified as follows: structured versus unstructured meshes, and body-fitted versus immersed boundary methods. 
Unstructured body-fitted meshes offer high geometric flexibility and are therefore well suited for complex configurations, but high-order methods are difficult to implement efficiently due to irregular cell connectivity and thus indirect memory access. 
Structured body-fitted meshes, such as curvilinear grids, are more favorable for high-order schemes because they retain regular cell connectivity, but complex geometries with fine geometric features are difficult to represent.

\begin{figure}[!t]
    \centering
    \includegraphics[width=\textwidth, trim=0.0cm 5cm 0.0cm 5cm, clip]{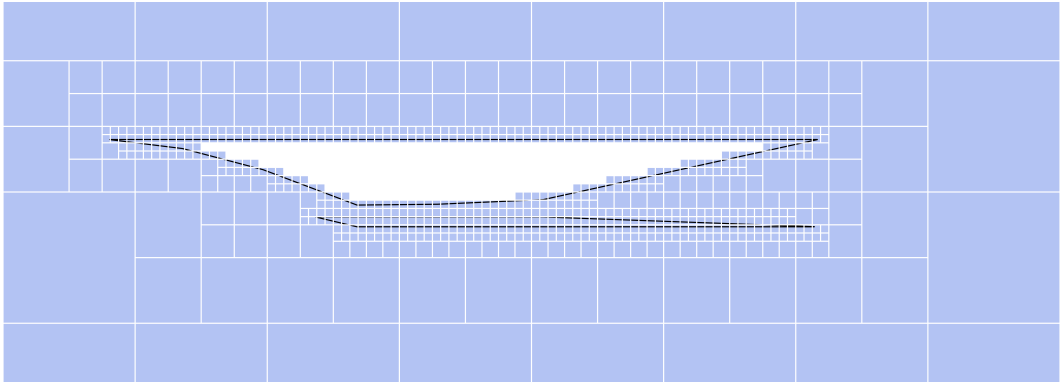}
    \caption{Schematic of the multi-block grid for the generic scramjet demonstrator configuration.}
    \label{fig:multiblock_mesh}
\end{figure}

In this work, we therefore employ a structured Cartesian multi-block mesh combined with a conservative cut-cell immersed-boundary method. 
The geometry is represented implicitly by a level-set function, and cells intersected by the fluid-solid interface are treated as cut cells. 
This approach combines the geometric flexibility of immersed boundary methods with the numerical efficiency of structured Cartesian grids. 
In particular, it enables automated meshing of fine geometric features while retaining the regular data layout and stencil structure required for efficient high-order finite-volume discretizations.
The multi-block formulation enables local refinement through quadtree- or octree-type subdivision, allowing high resolution near solid boundaries and interfaces while keeping a coarser resolution elsewhere. 
\cref{fig:multiblock_mesh} shows the multi-block grid of a representative geometry considered in this work. 

The Cartesian multi-block grid is particularly advantageous for GPU-accelerated simulation.
The solution variables within each block are stored as dense multidimensional arrays, leading to regular memory access, efficient vectorization, and reduced indirect addressing compared with unstructured grids.
As a result, the dominant numerical kernels are well matched to modern accelerators, where performance depends strongly on data locality, coalesced memory access, and high arithmetic throughput.

\paragraph{GPU-acceleration.} JAX-Fluids is implemented in JAX and compiled through XLA for execution on modern accelerators.
Within each block, the solver operations are
expressed as batched array operations, enabling large numbers of cells to be processed in parallel.
The multi-block decomposition also provides a natural parallelization strategy: a set of individual blocks can be assigned to different XLA devices, while communication is limited to halo exchanges.
In JAX-Fluids, this is enabled through JAX primitives such as
\texttt{jax.shard\_map} and \texttt{jax.lax.ppermute}.
This GPU-resident design is essential for generating large datasets over broad parameter spaces, including geometry and Mach number.
It is also important when JAX-Fluids is used inside the training loop to evaluate physics-based losses.

\paragraph{Differentiable solver.} JAX-Fluids is a fully differentiable solver that allows calculation of gradients of objective functions by automatic differentiation. 
These gradients are consistent with the discretized PDEs, i.e., they are consistent with the governing equation and the chosen numerical discretization.
Flow field predictions of a trained PE can be passed through JAX-Fluids to fine-tune it for achieving high-fidelity, physically-aware predictions in complex flows without inconsistencies with the data-generating numerics.

\subsection{Neural Architectures for Physics Emulators}

We investigate two neural architectures and two different modeling paradigms, each offering distinct trade-offs for predicting hypersonic flowfields on octree-based Euclidean mesh data. 
The octree structure decomposes the domain into axis-aligned blocks of uniform resolution, i.e., the count and ordering of blocks vary between cases.
This meshing strategy is increasingly adopted by modern GPU-based solvers because it maps naturally to parallel hardware \citep{jaber_gpunative_2026,carreon2025gpubasedcompressiblecombustionsolver}, making scalable emulation of such grids a broadly relevant objective.
To enable regular-grid architectures to be invariant to block count and ordering, we encode each patch's physical coordinates using complementary absolute (sinusoidal) and relative (rotary) positional encodings.
Combined, they give a single global attention a complete picture of patch location and pairwise displacement across blocks.
In contrast, architectures based on point-wise representations can simply ingest the raw point cloud.
This enables a range of neural architectures, and we consider the following instantiations of the PE:
\begin{enumerate*}[label=(\roman*)]
    \item a field-based approach designed for irregular grids \citep[AB-UPT]{alkin2025ab},
    \item a regular-grid-based vision transformer \citep[ViT]{dosovitsky2021vit},
    \item a generative flow matching model operating on the regular-grid representation \citep{lipman2022flow}.
\end{enumerate*}

\cref{tab:feature_matrix} summarizes the qualitative trade-offs among the three approaches.
AB-UPT treats the mesh as an irregular point cloud and yields smooth predictions but does not have built-in uncertainty estimates.
The ViT, by contrast, ingests the regular-grid octree representation and benefits from the highly optimized attention kernels available for uniform tensor data.
The flow matching model is also based on a regular-grid, but replaces the deterministic prediction head with a stochastic denoising process, trading single-pass efficiency for predictive uncertainty estimates.

Each model takes as input the simulation grid, either as a point cloud (AB-UPT) or in the block-stacked regular-grid representation (ViT, Flow Matching), together with a conditioning vector comprising the 15 geometry parameters of the scramjet configuration and the free-stream Mach number $M_\infty$.
The neural PE then predicts the corresponding flow field as a multi-channel output with C=8 channels, where each channel represents a different physical quantity, namely pressure~$p$, density~$\rho$, velocity~$\mathbf{u}$, temperature~$T$, enthalpy~$h$, total pressure~$P_t$, kinetic energy~$k$, and Mach number~$\operatorname{Ma}$.
The model is trained by minimizing the mean squared error over all grid points,
\begin{equation}
    \mathcal{L}_\text{data} = \frac{1}{N} \sum_{i=1}^{N} \left\| \hat{\mathbf{y}}_i - \mathbf{y}_i \right\|^2,
\end{equation}
where $\mathbf{y}_i \in \mathbb{R}^C$ is the ground-truth flow state at grid point $i$, $\hat{\mathbf{y}}_i$ is the corresponding prediction, and $N$
is the number of grid points.
Predicting primitive (i.e., density, velocity, and pressure) and derived quantities (i.e., temperature, enthalpy, total pressure, kinetic energy, and Mach number) jointly avoids the error accumulation that arises when derived fields are reconstructed from predicted primitives (see \cref{app:thermodynamic_consistency}).
The trade-off is that the predicted derived quantities are not guaranteed to be consistent with those recomputed from the predicted primitives. 
However, supervising only the primitives would leave the derived fields without a direct training signal and folding their computation into the loss introduces potentially ill-conditioned gradients through the nonlinear derivations as well as additional weighting hyperparameters across heterogeneous scales.

\begin{table}[t]
  \centering
  \caption{Qualitative comparison of neural physics emulator instantiations. 
           \cmark\;= fully satisfied, 
           \xmark\;= not satisfied.
           Different methods and training paradigms exhibit different advantages.}
  \vspace{.5em}
  \label{tab:feature_matrix}
  \renewcommand{\arraystretch}{1.25}
  \setlength{\tabcolsep}{6pt}
  \newlength{\colw}
  \setlength{\colw}{2.4cm}
  \begin{tabular}{@{} l *{3}{>{\centering\arraybackslash}p{\colw}} @{}}
    \toprule
    \multirow{2}{*}{\textbf{Characteristic}} & \textbf{AB-UPT} & \textbf{ViT}  & \textbf{Flow Matching} \\
    & \citep{alkin2025ab} & \citep{dosovitsky2021vit} & \citep{lipman2022flow} \\
    \midrule
    Point cloud         & \cellg\cmark & \cellr\xmark & \cellr\xmark \\
    Block-stacked regular grid     & \cellr\xmark & \cellg\cmark & \cellg\cmark \\
    Native UQ support                 & \cellr\xmark & \cellr\xmark & \cellg\cmark \\
    Single-pass inference                 & \cellg\cmark & \cellg\cmark & \cellr\xmark \\
    \bottomrule
  \end{tabular}
\end{table}

\subsection{Physics-aware Model Refinement}
The third phase of our workflow refines the pre-trained base model through physics-aware optimization to facilitate physical consistency.
This stage exploits the end-to-end differentiability of the JAX-Fluids solver to embed the governing flow equations directly within the ML optimization loop.
Whereas the initial training phase relies on a standard supervised loss the refinement phase minimizes the point-wise discrete residuals of the compressible Euler equations
\begin{equation}
   \mathcal{L}_\text{PDE} = \sum_{(i,j) \in \Omega}
    \sum_{k=1}^{4} w_k \left( R^k_{i,j} \right)^{2} \Delta x \Delta y,
\end{equation}
where $R^k_{i,j}$ denotes the point-wise residual for the $k$-th conserved quantity in cell $(i,j)$. $w_k$ is the corresponding loss weight. Here, the conserved quantities comprise mass, x-momentum, y-momentum, and total energy.
For details we refer to \cref{app:physics-loss}.

Using JAX-Fluids, we evaluate the residual-based loss $\mathcal{L}_\text{PDE}$ on the simulation mesh with the same numerical discretizations employed during data generation, ensuring consistency between training and inference.
A key advantage of this approach is that it improves generalization and can be performed in a target-free manner. 
It requires only a differentiable numerical solver and simulation meshes for the configurations of interest, i.e., no additional ground-truth solutions are needed. 
The configuration space can therefore be cheaply expanded beyond that of the training data since the loss is computed directly from the flow equations. 
This makes our model refinement a comparatively inexpensive fine-tuning procedure that avoids the cost of running new high-fidelity simulations.

\section{Experiments}
\label{sec:experiments}
This section details our experimental findings. 
We begin with an overview of the two datasets utilized for model training. 
We then present the training details and results of the baseline model, contrasting the various methodological approaches introduced previously. 
Within this analysis, we focus on scalability and provide results for scaling along the data and model axis.
Finally, we present the results of the model refinement phase. 

\subsection{Datasets}
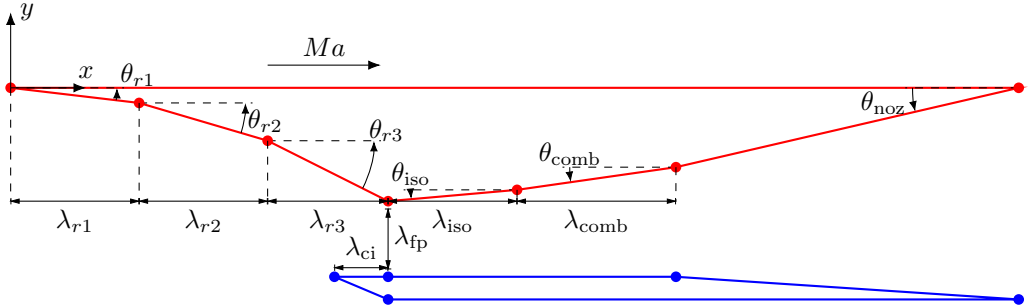
\begin{figure}[b]
    \centering
    \begin{tikzpicture}[
    scale=10,
    >=Latex,
    annot/.style={
        fill=white,
        inner sep=1pt
    },
    angguide/.style={
        black,
        dashed
    },
    anglearc/.style={
        draw=black,
        thin,
        -{Latex[length=4pt]},
        angle radius=40pt,
        angle eccentricity=1.2,
    },
    dimlabel/.style={
        fill=white,
        inner sep=1pt,
        sloped=false
    }
]

\pgfmathsetmacro{\xZero}{0}
\pgfmathsetmacro{\xA}{0.42838}
\pgfmathsetmacro{\xB}{0.4994}
\pgfmathsetmacro{\xC}{0.8802}
\pgfmathsetmacro{\xD}{1.3337}

\pgfmathsetmacro{\AyOne}{-0.25}
\pgfmathsetmacro{\AyTwo}{-0.28}

\pgfmathsetmacro{\xrefLen}{0.15}

\coordinate (A0) at (\xA,\AyOne);
\coordinate (A1) at (\xB,\AyOne);
\coordinate (A4) at (\xC,\AyOne);
\coordinate (A5) at (\xD,\AyTwo);
\coordinate (A9) at (\xB,\AyTwo);

\coordinate (B0) at (\xZero,0);
\coordinate (B1) at (0.17,-0.02);
\coordinate (B2) at (0.34,-0.07);
\coordinate (B3) at (\xB,-0.15);
\coordinate (B4) at (0.67,-0.135);
\coordinate (B5) at (\xC,-0.105);
\coordinate (B6) at (\xD,0);

\draw[thick,blue]
    (A0)--(A1)--(A4)--(A5)--(A9)--cycle;

\draw[thick,red]
    (B0)--(B1)--(B2)--(B3)--(B4)--(B5)--(B6)--cycle;

\draw[->] ([yshift=0.1cm]B2) -- ([yshift=0.1cm,xshift=0.15cm]B2) node[midway, above] {$Ma$};

\coordinate (B0x) at ($(B0)+(\xrefLen,0)$);
\coordinate (B1x) at ($(B1)+(\xrefLen,0)$);
\coordinate (B2x) at ($(B2)+(\xrefLen,0)$);

\coordinate (B4xBack) at ($(B4)-(\xrefLen,0)$);
\coordinate (B5xBack) at ($(B5)-(\xrefLen,0)$);
\coordinate (B6xBack) at ($(B6)-(\xrefLen,0)$);

\draw[angguide] (B0) -- (B0x);
\draw[angguide] (B1) -- (B1x);
\draw[angguide] (B2) -- (B2x);

\draw[angguide] (B4) -- (B4xBack);
\draw[angguide] (B5) -- (B5xBack);
\draw[angguide] (B6) -- (B6xBack);

\pic[anglearc, "$\theta_{r1}$" above]
    {angle = B1--B0--B0x};

\pic[anglearc, "$\theta_{r2}$"]
    {angle = B2--B1--B1x};

\pic[anglearc, "$\theta_{r3}$" {xshift=-2pt, yshift=15pt}]
    {angle = B3--B2--B2x};

\pic[
    anglearc,
    angle eccentricity=1.0,
    "$\theta_{\mathrm{iso}}$" above
]
    {angle = B4xBack--B4--B3};

\pic[
    anglearc,
    angle eccentricity=1.0,
    "$\theta_{\mathrm{comb}}$" above
]
    {angle = B5xBack--B5--B4};

\pic[
    anglearc,
    angle eccentricity=1.28,
    "$\theta_{\mathrm{noz}}$"
]
    {angle = B6xBack--B6--B5};

\foreach \p in {A0,A1,A4,A5,A9}
    \fill[blue] (\p) circle (0.2pt);

\foreach \p in {B0,B1,B2,B3,B4,B5,B6}
    \fill[red] (\p) circle (0.2pt);

\draw[->] (B0) -- ($(B0)+(0.1,0)$) node[at end, above] {$x$};
\draw[->] (B0) -- ($(B0)+(0,0.1)$) node[at end, right] {$y$};

\coordinate (B0hDim) at (B0 |- B3);
\coordinate (B1hDim) at (B1 |- B3);
\coordinate (B2hDim) at (B2 |- B3);
\coordinate (B3hDim) at (B3);
\coordinate (B4hDim) at (B4 |- B3);
\coordinate (B5hDim) at (B5 |- B3);

\draw[angguide] (B0) -- (B0hDim);
\draw[angguide] (B1) -- (B1hDim);
\draw[angguide] (B2) -- (B2hDim);
\draw[angguide] (B4) -- (B4hDim);
\draw[angguide] (B5) -- (B5hDim);

\dimline[
    extension start length=0cm,
    extension end length=0cm,
    line style={thin},
    label style={dimlabel, below=2pt}
]
{(B0hDim)}
{(B1hDim)}
{$\lambda_{r1}$};

\dimline[
    extension start length=0cm,
    extension end length=0cm,
    line style={thin},
    label style={dimlabel, below=2pt}
]
{(B1hDim)}
{(B2hDim)}
{$\lambda_{r2}$};

\dimline[
    extension start length=0cm,
    extension end length=0cm,
    line style={thin},
    label style={dimlabel, below=2pt}
]
{(B2hDim)}
{(B3hDim)}
{$\lambda_{r3}$};

\dimline[
    extension start length=0cm,
    extension end length=0cm,
    line style={thin},
    label style={dimlabel, below=2pt}
]
{(B3hDim)}
{(B4hDim)}
{$\lambda_{\mathrm{iso}}$};

\dimline[
    extension start length=0cm,
    extension end length=0cm,
    line style={thin},
    label style={dimlabel, below=2pt}
]
{(B4hDim)}
{(B5hDim)}
{$\lambda_{\mathrm{comb}}$};

\dimline[
    extension start length=0cm,
    extension end length=0cm,
    line style={thin},
    label style={dimlabel, right=1pt}
]
{($(B3)-(0,0.010)$)}
{($(A1)+(0,0.010)$)}
{$\lambda_{\mathrm{fp}}$};

\dimline[
    extension start length=0cm,
    extension end length=0cm,
    line style={thin},
    label style={dimlabel, above=2pt}
]
{($(A0)+(0,0.012)$)}
{($(A1)+(0,0.012)$)}
{$\lambda_{\mathrm{ci}}$};

\end{tikzpicture}
    \caption{
    \textbf{Schematic of the parametrized scramjet geometry illustrating the design-parameters} $\mathbf{p}$.
    Here, $\lambda_{r1},\lambda_{r2},\lambda_{r3}$ are the intake ramp length fractions. $\lambda_{\mathrm{iso}}$ and $\lambda_{\mathrm{comb}}$ denote the isolator and combustor length fractions. $\theta_{r1},\theta_{r2},\theta_{r3}$ are the intake ramp angles, and $\theta_{\mathrm{iso}},\theta_{\mathrm{comb}},\theta_{\mathrm{noz}}$ the isolator, combustor, and nozzle angles, all measured relative to the $x$-axis. $\lambda_{\mathrm{fp}}$ is the flow path height fraction and $\lambda_{\mathrm{ci}}$ is the cowl intake length fraction. The length fractions are defined with respect to a fixed reference length and the cowl has a fixed height. In addition to the geometry parameters, the design parameters $\mathbf{p}$ also contains the inflow Mach number $Ma$.
    }
    \label{fig:geometry_params}
\end{figure}

The experimental evaluation utilizes two distinct datasets of scramjet configurations in hypersonic conditions, $D1$ and $D2$, both generated using JAX-Fluids as detailed in \cref{subsec:jax-fluids}. 
Each scramjet case is parameterized by its design-parameter vector $\mathbf{p}_i \in \mathbb{R}^d$ (see Figure~\ref{fig:geometry_params}), where $\mathbf{p}_i$ concatenates geometry and inflow parameters of case $i$. $D1$ represents a high-fidelity dataset, whereas $D2$ is target-free, i.e., it only comes with mesh and input parameters without ground-truth field data.
For the pre-training stage of our models, we rely on $D1$, whereas for fine-tuning we also leverage $D2$.
We list both datasets and their corresponding properties in \cref{tab:datasets}.

\begin{table}[h]
    \centering
    \caption{Overview of the two datasets used for model training and evaluation. $D_1$ prioritizes numerical accuracy through minimized residuals, while $D_2$ comprises only meshes for new scramject parameterizations.}
    \vspace{.5em}
    \label{tab:datasets}
    \begin{tabular}{lcc}
        \toprule
        \textbf{Property} & \textbf{D1} & \textbf{D2} \\
        \midrule
        Parameters & \cmark & \cmark \\
        Mesh & \cmark & \cmark \\
        Field-data & \cmark & \xmark \\
        Generation time  & Steady-state simulation            & Only meshing            \\
        Number of samples        & $7081$           & $7845$            \\
        Primary use              & Accuracy & Refinement \\
        \bottomrule
    \end{tabular}
\end{table}

To evaluate the different neural PEs we construct different dataset splits.
Stacking the design-parameter vectors of all cases considered yields the parameter matrix $\mathbf{P} \in \mathbb{R}^{N \times d}$.
First, we fit an Isolation Forest \citep{liu_2008_isolationforest} with $T = 100$ trees on the parameter matrix $\mathbf{P}$ and obtain an anomaly score $s_i = \operatorname{decision\_function}(\mathbf{p}_i)$ for each case in the dataset, where $s_i$ corresponds to points that are isolated by shorter random partitioning paths and are therefore marked anomalous in design space. 
We select 10\% of cases with the lowest scores and assemble them into an out-of-distribution (OOD) set.
The remaining in-distribution pool is then randomly partitioned into train/val/test according to a 80/10/10 split. 
We verify the construction post-hoc by visualizing the anomaly scores of the Isolation Forest per split (see \cref{fig:ood_splits} in \cref{app:data_gen}), confirming that OOD samples lie on the periphery of the parameter manifold while train/val/test overlap in the interior.

\subsection{Evaluation Protocol}

We evaluate each PE against the reference JAX-Fluids simulations on three engineering key performance indicators (KPIs) that quantify aspects of scramjet performance, supplemented by a qualitative comparison of the density field. 
They are derived from the full predicted flow, so that they expose the downstream consequences of the predictions, such as total-pressure recovery up to the inlet to combustor, cumulative total-pressure loss across the scramjet from the inlet plane to the nozzle exit, and peak thermal load on the wetted geometry.

\paragraph{Total-pressure ratio $\pi_d$.} 
Following standard inlet-performance practice, we define
\begin{equation}
\pi_d = \frac{ \left< p_t\vert_{x_\mathrm{stn2}} \right> } { \left< p_t\vert_{x_\mathrm{inlet}} \right> }     
\end{equation}
i.e., the ratio of mass-flux-weighted total pressure between the inlet plane and station 2, taken at the axial mid-point of the scramjet and restricted to the fluid side of the embedded geometry by the signed-distance level set. 
An ideal isentropic compression delivers $\pi_d=1$, whereas losses (e.g., due to shocks) between capture surface and entrance of the combustor reduce $\pi_d$.

\paragraph{Total-pressure loss ratio $\Lambda$.} 
Whereas $\pi_d$ characterizes the inlet up to station 2, the total-pressure loss ratio
\begin{equation}
\Lambda = \frac{ \left< p_t\vert_{x_\mathrm{inlet} } \right> - \left< p_t\vert_{x_\mathrm{exit}} \right> } { \left< p_t\vert_{x_\mathrm{inlet}} \right> }
\end{equation}
captures the cumulative loss across the entire duct, from the inlet plane to the nozzle exit. 
The two pressure-based KPIs are deliberately complementary.

\paragraph{Peak surface temperature.} 
The third KPI is a structural-design figure of merit, namely the peak temperature on the scramjet wetted surface. 
From the predicted temperature fields we extract the peak surface temperature and report the 95th percentile. 
Peak wall temperature in a hypersonic intake drives the choice of thermal-protection material and the cooling-system budget for the vehicle.

\begin{table}[tb!]
\centering
\caption{
\textbf{Performance of different physics emulators on hypersonic flowfields.}
Relative L2 errors (\%) for pressure $p$, density $\rho$, velocity $\bm{u}$, enthalpy $h$, total pressure $p_t$, kinetic energy $k$, temperature $T$, and Mach number $\operatorname{Ma}$ for AB-UPT, ViT, and Flow Matching (FM) across three random seeds.}
\vspace{.5em}
\label{tab:rel_err_flowfields}
\setlength{\tabcolsep}{6pt}

\resizebox{\textwidth}{!}{
\begin{tabular}{llcccccccc}
\toprule
\textbf{Split} & \textbf{Model} & $p$ &
$\rho$ & $\bm{u}$ & $h$ & $p_t$ & $k$ & $T$ & $\operatorname{Ma}$ \\
\midrule

\multirow{3}{*}{Val}
& AB-UPT    & $2.16_{\pm 0.015}$ & $1.77_{\pm 0.013}$ & $0.38_{\pm 0.000}$ & $1.42_{\pm 0.001}$ & $2.33_{\pm 0.028}$ & $1.76_{\pm 0.019}$ & $1.42_{\pm 0.001}$ & $0.81_{\pm 0.006}$ \\
& ViT      & $2.86_{\pm 0.017}$ & $2.24_{\pm 0.015}$ & $0.48_{\pm 0.005}$ & $1.65_{\pm 0.021}$ & $2.62_{\pm 0.043}$ & $2.15_{\pm 0.018}$ & $1.65_{\pm 0.021}$ & $0.95_{\pm 0.015}$ \\
& FM & $8.64_{\pm 1.211}$ & $6.15_{\pm 0.787}$ & $1.13_{\pm 0.099}$ & $3.13_{\pm 0.202}$ & $3.65_{\pm 0.144}$ & $5.03_{\pm 0.536}$ & $3.13_{\pm 0.202}$ & $1.55_{\pm 0.053}$ \\
\midrule
\multirow{3}{*}{Test}
& AB-UPT   & $2.14_{\pm 0.018}$ & $1.76_{\pm 0.014}$ & $0.38_{\pm 0.001}$ & $1.43_{\pm 0.009}$ & $2.33_{\pm 0.008}$ & $1.75_{\pm 0.022}$ & $1.43_{\pm 0.009}$ & $0.82_{\pm 0.005}$ \\
& ViT   &   $2.84_{\pm 0.038}$ & $2.24_{\pm 0.023}$ & $0.48_{\pm 0.005}$ & $1.68_{\pm 0.014}$ & $2.63_{\pm 0.021}$ & $2.16_{\pm 0.016}$ & $1.68_{\pm 0.014}$ & $0.96_{\pm 0.006}$ \\
& FM & $8.51_{\pm 1.263}$ & $6.06_{\pm 0.831}$ & $1.12_{\pm 0.113}$ & $3.12_{\pm 0.222}$ & $3.68_{\pm 0.111}$ & $4.97_{\pm 0.559}$ & $3.12_{\pm 0.222}$ & $1.56_{\pm 0.051}$  \\
\midrule
\multirow{3}{*}{OOD}
& AB-UPT   & $3.13_{\pm 0.038}$ & $2.41_{\pm 0.042}$ & $0.51_{\pm 0.010}$ & $1.78_{\pm 0.028}$ & $2.77_{\pm 0.058}$ & $2.23_{\pm 0.046}$ & $1.78_{\pm 0.028}$ & $0.99_{\pm 0.008}$ \\
& ViT      & $4.01_{\pm 0.054}$ & $2.99_{\pm 0.043}$ & $0.62_{\pm 0.008}$ & $2.03_{\pm 0.026}$ & $2.95_{\pm 0.022}$ & $2.69_{\pm 0.029}$ & $2.03_{\pm 0.026}$ & $1.13_{\pm 0.011}$ \\
& FM & $9.89_{\pm 1.375}$ & $6.97_{\pm 0.892}$ & $1.28_{\pm 0.112}$ & $3.55_{\pm 0.225}$ & $4.11_{\pm 0.123}$ & $5.63_{\pm 0.565}$ & $3.55_{\pm 0.225}$ & $1.75_{\pm 0.043}$ \\
\bottomrule

\end{tabular}}
\end{table}

\subsection{Pre-Training Results}

We present results for the three different instantiations of PEs and their performance in the following sections.
All of the results in this section are exclusively based on training on $\mathcal{L}_\text{data}$ and cover a two different architectures and modeling paradigms within a typical toolbox for providing PE solutions for engineering problems.

\subsubsection{AB-UPT}
\begin{figure}[ht]
  \centering
  \includegraphics[width=\linewidth]{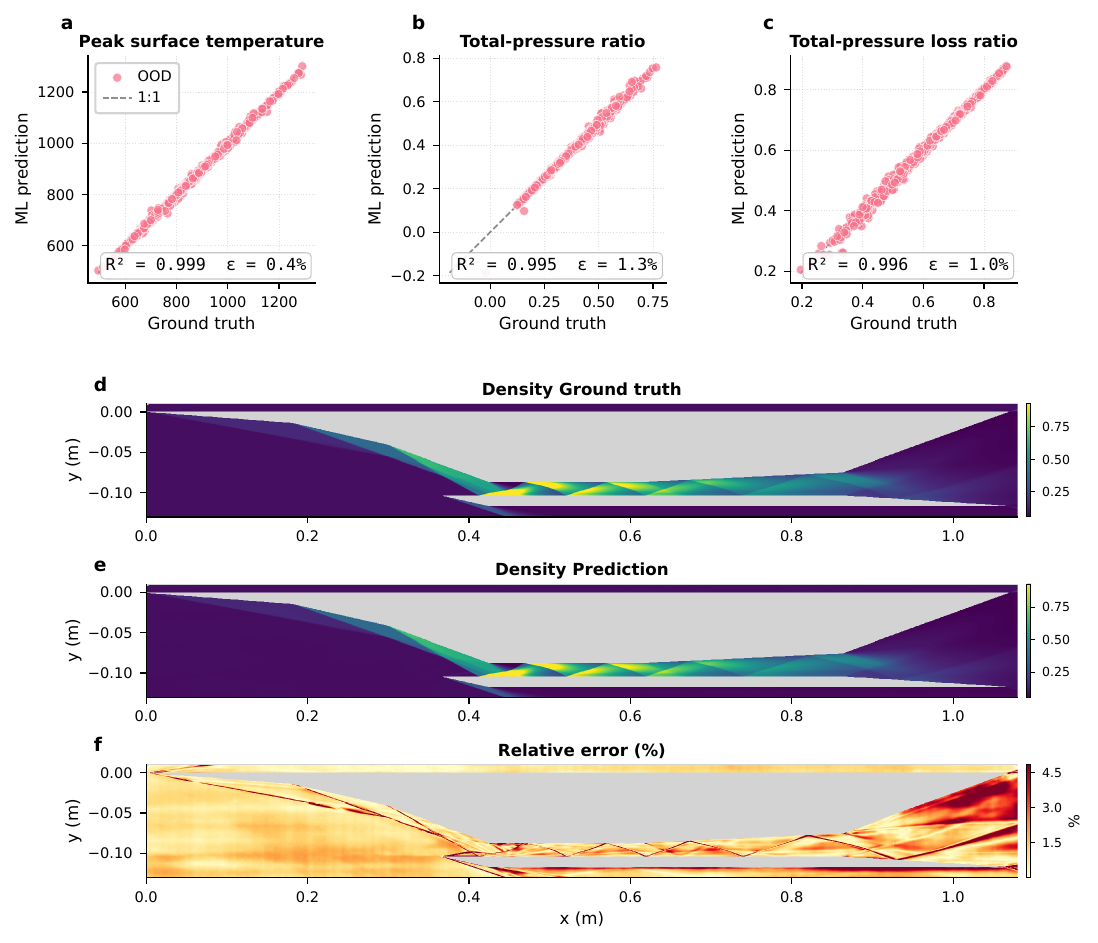}
  \caption{\textbf{Prediction error of AB-UPT on hypersonic scramjet.}
  Parity plots of predicted vs ground-truth on the OOD test data for \textbf{a},~peak surface temperature, \textbf{b},~total pressure ratio, and \textbf{c},~total pressure loss.
  We also report the ground-truth \textbf{d} and predicted \textbf{e} density field along with the relative error in percentage points \textbf{f} for a random case in the OOD test data.
  }
  \label{fig:abupt_combined_plot}
\end{figure}

As illustrated in \cref{fig:abupt_combined_plot}, the model maintains high fidelity across the relevant metrics. 
Parity plots for $\pi_d$, $\Lambda$, and peak surface temperature (\cref{fig:abupt_combined_plot}a-c) show tight clustering along the identity line.
Furthermore, the density field predictions (\cref{fig:abupt_combined_plot}e) closely mirror the ground-truth solver (\cref{fig:abupt_combined_plot}d). 
The relative error map (\cref{fig:abupt_combined_plot}f) reveals that the highest discrepancies are localized at shockwaves and the exhaust nozzle with an average relative $L_2$ error around 3\%.
Finally, we report performance metrics in \cref{tab:performance_metrics} showing that training and inference speed for AB-UPT is slower than other methods.
Specifically, training sequences are longer due to the selection of anchor points and inference speed is much slower as AB-UPT decodes field values for each grid position during a single forward pass.
This also results in elevated GFLOPs and memory footprint.

\begin{figure}
    \centering
    \includegraphics[width=\linewidth]{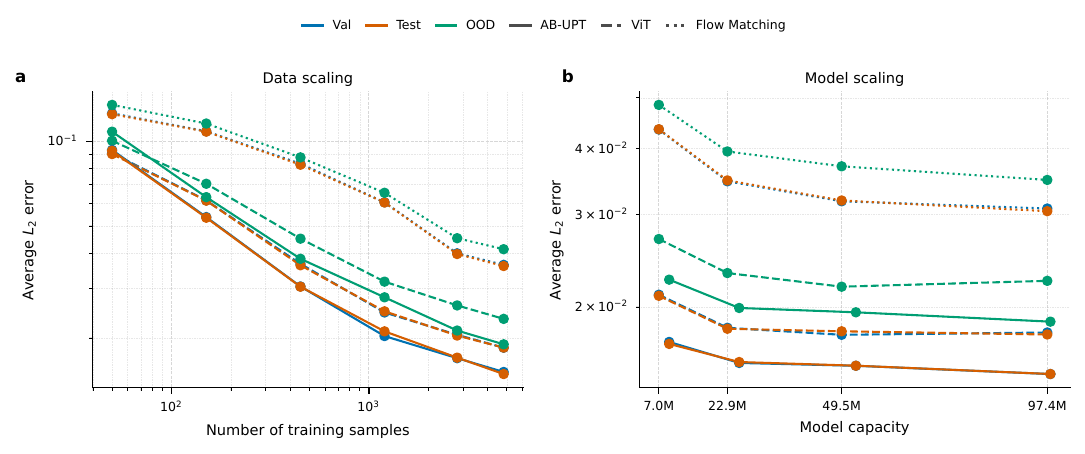}
    \caption{\textbf{Scaling laws for the different modeling paradigms.} Data \textbf{(a)} and model scaling laws \textbf{(b)} for AB-UPT, ViT, and Flow Matching on validation, test, and OOD extrapolation sets.}
    \label{fig:data_model_scale}
\end{figure}

Additionally, we analyze data and model scaling as shown in \cref{fig:data_model_scale}. 
For data scaling, we train a model with $\sim 25M$ parameters on subsampled training splits.
The different splits comprise $\{50,150,450,1200,2800,4816\}$ samples using farthest point sampling to maintain coverage of the design parameter space.
For training details and hyperparameter searches we refer the reader to \cref{app:scaling}.
For model scaling experiments, we scale the model from $\sim 9M$ parameters to $\sim 100M$ parameters.
AB-UPT exhibits a steady trend of improvement for data scaling on validation, test, and OOD splits and achieves the lowest loss on all datasets compared to competitors.
The trend for model scaling differs slightly.
There is a noticeable improvement when scaling the model from $\sim 9M$ to $\sim 25M$ parameters.
Beyond $25M$ parameters the test error flattens and further increases in depth yield marginal improvements. 
We interpret this as a data-bottlenecked regime in which the model's representational capacity already exceeds what the training distribution can support.

\subsubsection{ViT}
\label{subsec:vit}

\begin{figure}[ht]
  \centering
  \includegraphics[width=\linewidth]{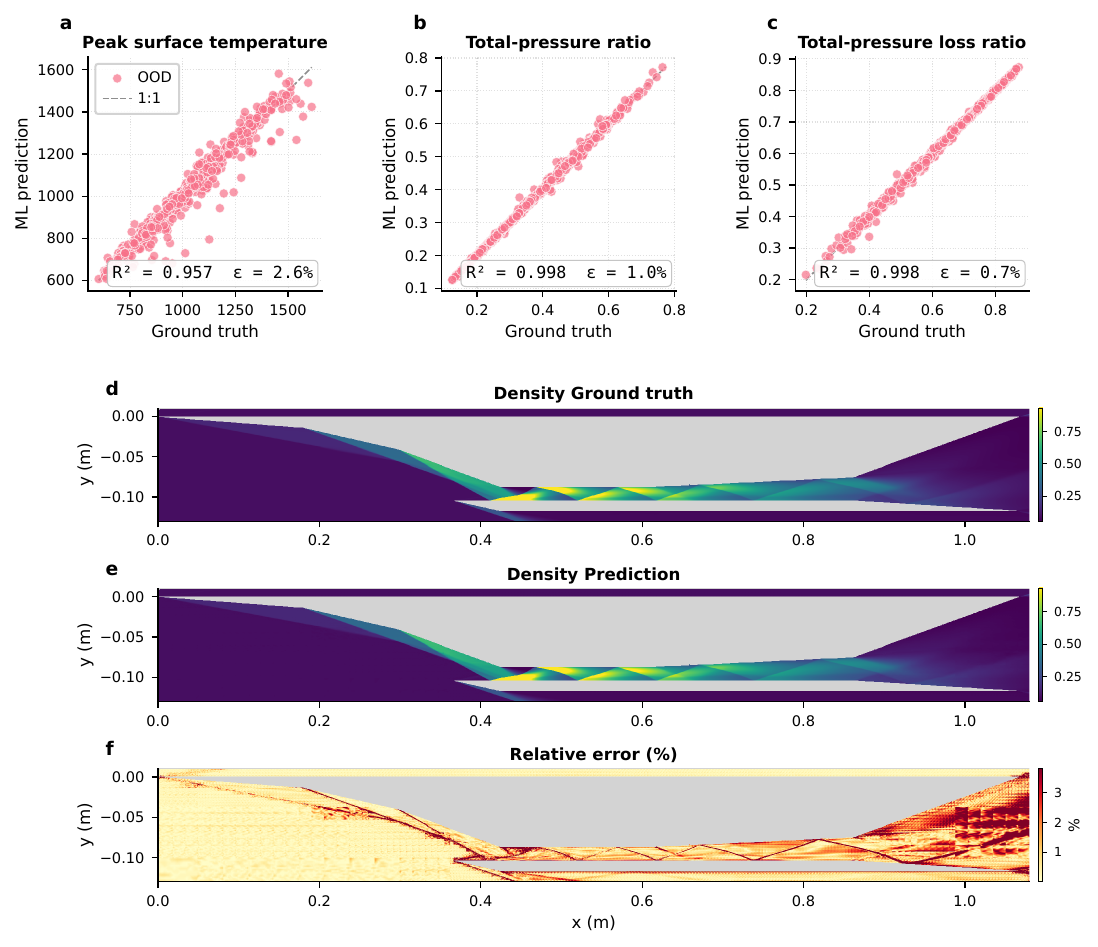}
  \caption{\textbf{Prediction error of ViT on hypersonic scramjet.}
  Parity plots of predicted vs ground-truth on the OOD test data for \textbf{a},~peak surface temperature, \textbf{b},~total pressure ratio, and \textbf{c},~total pressure loss.
  We also report the ground-truth \textbf{d} and predicted \textbf{e} density field along with the relative error in percentage points \textbf{f} for a random case in the OOD test data.
  }
  \label{fig:vit_combined_plot}
\end{figure}

We present the results for our trained ViT model in \cref{fig:vit_combined_plot}.
The KPIs are comparable to the ones achieved by AB-UPT.
Further, the error distribution shows higher error around shockwaves, expansion fans, reflected shocks,  and in the exhaust nozzle.
Especially at the exhaust nozzle the error pattern is different to the one obtained by AB-UPT.
Specifically, the error pattern appears to be an artifact of patching.
A potential remedy would be reducing the patch size or adding a smoothing operation (e.g., by convolution) in the decoder.
We experimented with the latter and found no significant performance gains.
Furthermore, decreasing the patch size leads to longer token sequences which results in substantial additional cost when performing self-attention as sequence length scales quadratically with patch size.
The performance metrics in \cref{tab:performance_metrics} clearly show that ViT is more efficient in terms of training and inference compared to AB-UPT while also reducing memory footprint and GFLOPs.

The data-scaling results in \cref{fig:data_model_scale} uncover an interesting finding.
At data-scarce regimes ($50$ samples) ViT attains lower average $L_2$ error across all fields compared to AB-UPT, however in larger data regimes AB-UPT recovers and outperforms ViT.
In the data-scarce regime, ViT's regular-grid structure provides a strong spatial inductive bias that reduces the number of learnable degrees of freedom.
AB-UPT must simultaneously infer both the spatial structure and the physical field mapping, a task for which $50$ training samples provide weak information.
In terms of model scale we observe similar behavior as for AB-UPT, again with a consistent offset in error.

\subsubsection{Flow Matching with ViT backbone}
\begin{figure}[ht]
  \centering
  \includegraphics[width=\linewidth]{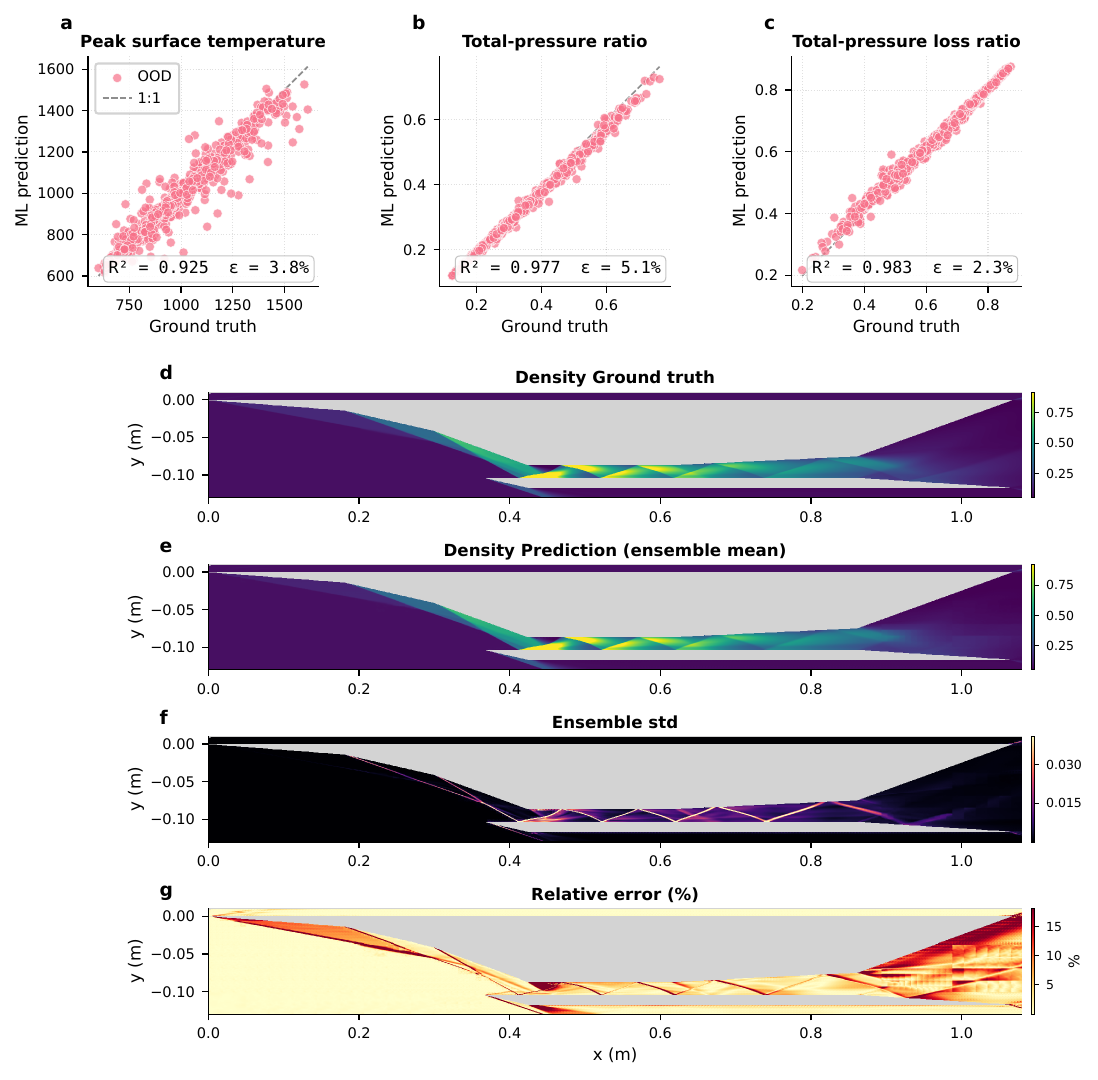}
  \caption{\textbf{Prediction error of Flow Matching on hypersonic scramjet.}
  Parity plots of predicted vs ground-truth on the OOD test data for \textbf{a},~peak surface temperature, \textbf{b},~total pressure ratio, and \textbf{c},~total pressure loss.
  We also report the ground-truth \textbf{d} and predicted \textbf{e} density field along with the ensemble standard deviation \textbf{f} and relative error in percentage points \textbf{g} for a random case in the OOD test data.
  }
  \label{fig:diffusion_combined_plot}
\end{figure}

We present the results for flow matching for an ensemble of 10 members in \cref{fig:diffusion_combined_plot}.
While we observe similar correlation coefficients for the different KPIs, the field error is significantly elevated for flow matching.
The predictive uncertainty clearly highlights that the highest uncertainty is located around shockwaves.
Compared to ViT predictions patching artifacts are visually less prevalent as we report the posterior mean over 10 ensemble members, effectively acting as a smoothing operator.
Considering performance metrics in \cref{tab:performance_metrics}, flow matching generally is efficient during training, but requires more memory and inference time due to ensembling compared to ViT.

The data-scale results in \cref{fig:data_model_scale} show that flow matching generally yields worse performance than AB-UPT or ViT.
However, we make several interesting observations.
First, the average gap between in-distribution and OOD evaluation splits is smaller for flow matching ($9.63\% \pm 3.69\%$) than for AB-UPT ($24.23\% \pm 5.72\%$) or ViT ($22.10\% \pm 7.11\%$).
Second, elevated test errors indicate that there appears to be a fundamental trade-off between accuracy and robustness for flow matching.
This might be traced back to the deterministic nature of the problem setup.
As flow matching is a probabilistic method it adds uncertainty estimation and additional robustness, at the cost of accuracy.
We observe similar model scaling behavior as for ViT, albeit with a consistent offset in error.

\begin{table}[tb!]
\centering
\caption{\textbf{Performance metrics for the different physics emulators.} We report number of parameters, latency, FLOPs, and peak memory consumption over 10 inference passes. AB-UPT processes the entire point cloud grid as queries and flow matching ensembles over 10 samples.}
\vspace{.5em}
\label{tab:performance_metrics}
\setlength{\tabcolsep}{6pt}

\resizebox{\textwidth}{!}{
\begin{tabular}{lccccc}
\toprule
\textbf{Model} & Params (M) & Train (ms) & Inference (ms) & FLOPs (G) &  %
Peak mem (MB) \\
\midrule
AB-UPT   & 25.67 & $29.64 \pm 0.13$ & $2089.87 \pm 22.95 $ & 506834.2 %
& 13739 \\
ViT & 22.92 & $10.06 \pm 0.02$ & $10.07 \pm 0.05$ & 757.9 %
&    282 \\
Flow Matching &  23.34 & $8.15 \pm 0.02$ & $237.54 \pm 6.08$ & 7602.1 %
& 375 \\
\bottomrule

\end{tabular}}
\end{table}

\begin{figure}[t]
    \centering
    \begin{subfigure}[t]{0.32\linewidth}
        \centering
        \includegraphics[width=\linewidth]{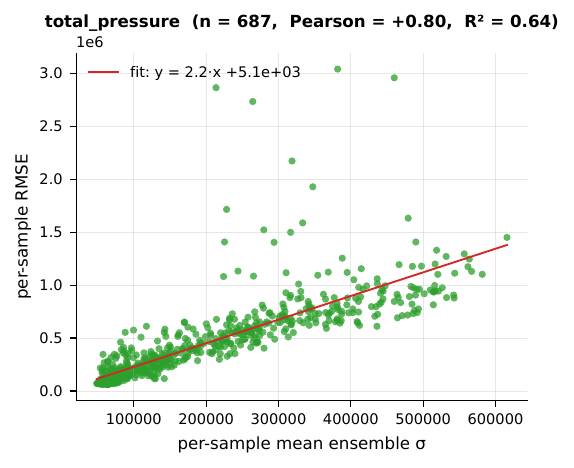}
        \caption{Flow matching}
        \label{fig:calib_pt_fm}
    \end{subfigure}\hfill
    \begin{subfigure}[t]{0.32\linewidth}
        \centering
        \includegraphics[width=\linewidth]{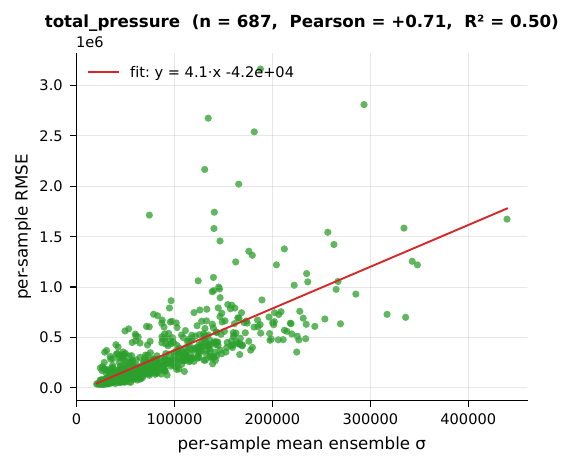}
        \caption{ViT}
        \label{fig:calib_pt_vit}
    \end{subfigure}\hfill
    \begin{subfigure}[t]{0.32\linewidth}
        \centering
        \includegraphics[width=\linewidth]{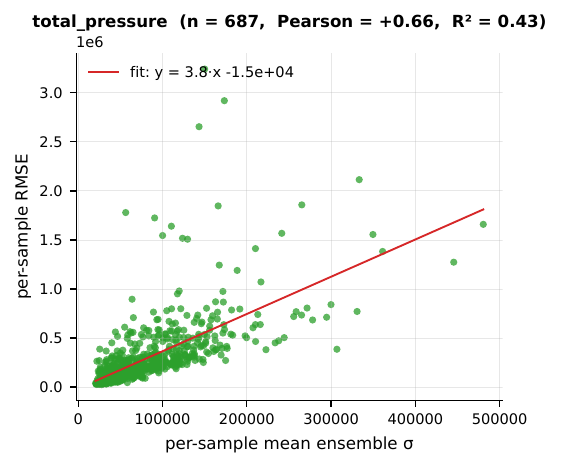}
        \caption{AB-UPT}
        \label{fig:calib_pt_abupt}
    \end{subfigure}
    \caption{\textbf{Correlation between predictive uncertainty and error of the total-pressure field across the three emulators.}
    Each panel compares the predicted distribution of $P_t$ against the JAX-Fluids
    reference on the out-of-distribution test split for the flow-matching ViT
    (\textbf{a}), the deterministic ViT (\textbf{b}) and the AB-UPT model
    (\textbf{c}). 
    We show $p_t$ as it is the direct integrand of the total-pressure ratio $\pi_d$ used as performance KPI. 
    Correlation of $p_t$ propagates one-to-one into $\pi_d$ at every probe location.
    The flow-matching surrogate is the most strongly calibrated on $p_t$. 
    }
    \label{fig:calibration_total_pressure}
\end{figure}

\textbf{Uncertainty Quantification}
In practice it is vital to obtain estimates on the predictive uncertainty of the emulator to inform decision making.
Regions that exhibit high predictive uncertainty indicate that model predictions may not be trusted.
We analyze the predictive uncertainty obtained by the different neural PEs. 
In particular, we visualize the correlation between the per-sample RMSE for cases in the OOD evaluation set and check for correlation with the per-sample mean of the ensemble variance.
To obtain an ensemble for the deterministic ViT and AB-UPT models, we train three model instances with different random seeds.
For flow matching we construct ensemble members of 10 initial noise samples and average them over three seeds.
We report our findings on the predicted total pressure field for flow matching (\cref{fig:calib_pt_fm}), ViT (\cref{fig:calib_pt_vit}), and AB-UPT (\cref{fig:calib_pt_abupt}).
Flow matching achieves the highest Pearson correlation ($0.80$) and coefficient of determination ($R^2=0.64$).
This pattern is consistent across other predicted fields such as enthalpy, kinetic energy, Mach number, and temperature as we show in \cref{fig:calibration_derived_fields}.

\subsection{Physics-aware Model Refinement}
\label{sec:model_refinement}
After pre-training, we perform physics-aware refinement aimed at improving physical consistency as well as predictive capabilities of the PE. 
Each refinement run starts from a pre-trained checkpoint and minimizes a weighted sum of up to three terms,
\begin{equation}
  \mathcal{L} = \mathcal{L}_\mathrm{data} + w_\mathrm{div}\,\mathcal{L}_\mathrm{div} + \lambda\,\mathcal{L}_\mathrm{PDE},
\end{equation}
namely a supervised data-reconstruction loss $\mathcal{L}_\mathrm{data}$, a divergence term from the base model $\mathcal{L}_\mathrm{div}$, and a physics loss $\mathcal{L}_\mathrm{PDE}$. 
$\mathcal{L}_\mathrm{div}$ and $\mathcal{L}_\mathrm{PDE}$ are target-free, i.e., they require only the mesh and input parameters, with no ground-truth field data. 
The physics loss exploits the differentiability of the JAX-Fluids solver, evaluating the point-wise residual of the discretized equations via the solver's residual operator (see Appendix~\ref{app:physics-loss}). 
Since neither target-free term consumes labeled data, refinement requires no additional simulations beyond those already used for pre-training.
We provide further implementation and training details in \cref{app:physics-loss}.

We present results for our proposed model refinement strategy showing that 
\begin{enumerate*}[label=(\roman*)]
    \item model refinement improves performance after pre-training, 
    \item the addition of the physics loss improves generalization, and
    \item refinement can be entirely target-free on simulation meshes without the need for solved flow-field data.
\end{enumerate*}
For this line of experiments we take the ViT model from \cref{subsec:vit} as a base model and conduct different fine-tuning experiments. 
The residual operator acts naturally on the ViT's regular grid, whereas recovering a full grid from AB-UPT is expensive (see \cref{tab:performance_metrics}).
Furthermore, ViT is the cheapest method in terms of FLOPs and inference speed compared to AB-UPT and flow matching,

To demonstrate the effectiveness of fine-tuning based on differentiating through the residual operator of JAX-Fluids, we perform the following experiments.
First, as a baseline we simply fine-tune the Base model for the same amount of steps as we use for other fine-tuning strategies (Base + Data).
Second, we add a loss term on the residual operator (Base + Data + Residual).
Finally, to show that no ground-truth data are required for the fine-tuning procedure, we remove the data loss term and train only on the residual loss, plus the divergence loss for regularization.
We include $\mathcal{L}_\mathrm{div}$ as only training on $\mathcal{L}_\mathrm{PDE}$ collapses to degenerate fields with near-zero residuals; the divergence term anchors the prediction to the pre-trained prior, supplying the missing constraint to select physically meaningful solutions.

\begin{figure}[tb!]
  \centering
  \begin{subfigure}[b]{\linewidth}
    \raggedright\textbf{a}\\[-.1em]
    \centering
    \includegraphics[width=\linewidth]{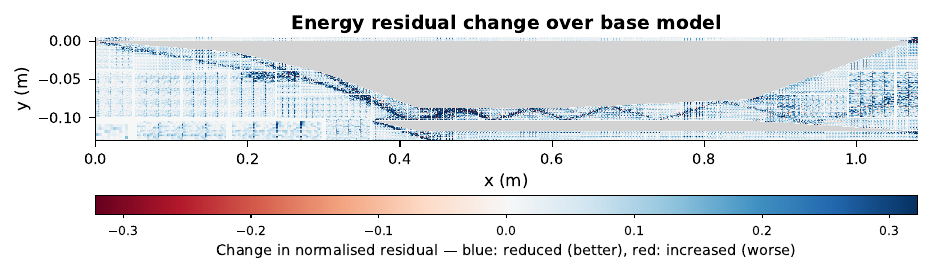}
    \phantomsubcaption\label{fig:panel_a}
  \end{subfigure}
  \\
  \begin{subfigure}[b]{\linewidth}
    \raggedright\textbf{b}\\[-.1em]
    \centering
    \includegraphics[width=\linewidth]{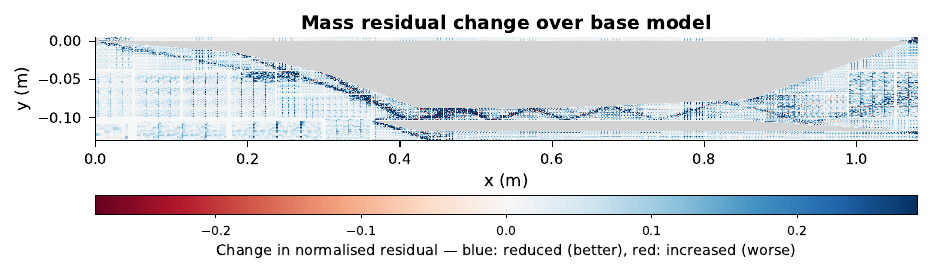}
    \phantomsubcaption\label{fig:panel_b}
  \end{subfigure}
  \\
  \begin{subfigure}[b]{\linewidth}
    \raggedright\textbf{c}\\[-.1em]
    \centering
    \includegraphics[width=\linewidth]{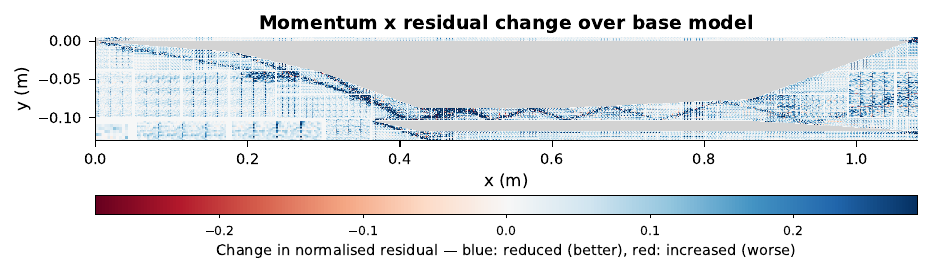}
    \phantomsubcaption\label{fig:panel_c}
  \end{subfigure}
  \\
  \begin{subfigure}[b]{\linewidth}
    \raggedright\textbf{d}\\[-.1em]
    \centering
    \includegraphics[width=\linewidth]{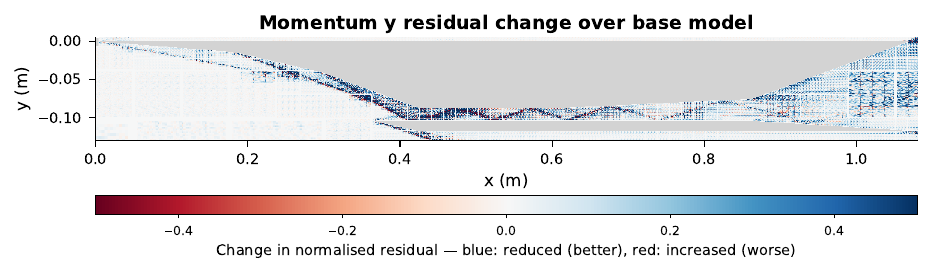}
    \phantomsubcaption\label{fig:panel_d}
  \end{subfigure}
  \caption{\textbf{Improvement of residuals after target-free refinement.}
  We report the change in normalized residuals where blue indicates improvement over the base model and red indicates worse residuals for
  \textbf{a},~energy conservation,
  \textbf{b},~mass conservation,
  \textbf{c},~x-component of momentum conservation, and
  \textbf{d},~y-component of momentum conservation.}
  \label{fig:pointwise_residual_per_term}
\end{figure}

\begin{figure}
    \centering
    \includegraphics[width=\linewidth]{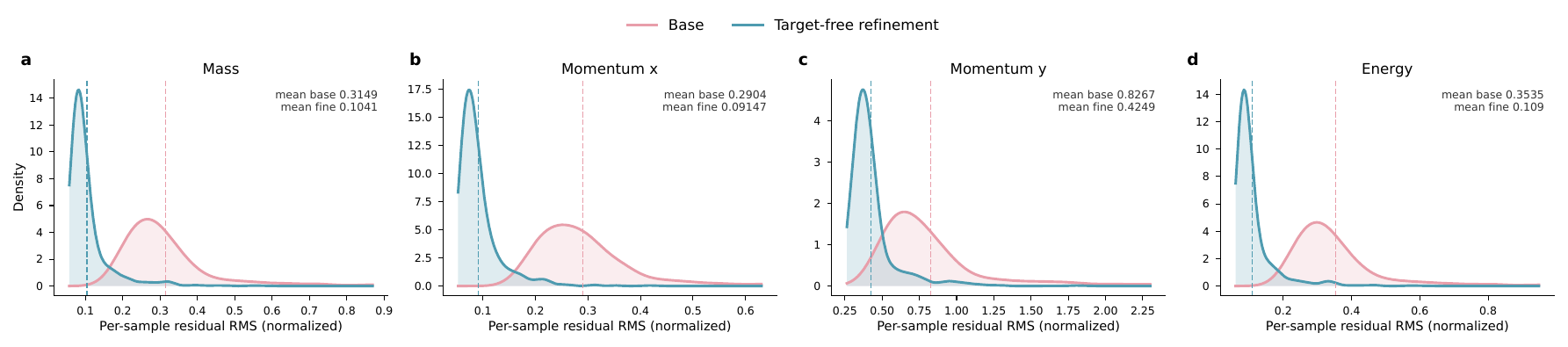}
    \caption{\textbf{Distribution of residuals on the OOD test set.} 
    We report residuals for the conservation terms mass \textbf{(a)}, momentum in x and y direction \textbf{(b--c)}, and energy \textbf{(d)}. 
    Base refers to the base model prior to refinement. 
    Target-free refinement significantly reduces residuals and hence improves physical consistency.
    }
    \label{fig:residual_dist_fine_tuning}
\end{figure}

We report the results on the different conservation terms in $\mathcal{L}_\text{PDE}$ in \cref{tab:rel_err_residuals_finetune}.
Our main findings are three fold, namely \begin{enumerate*}[label=(\roman*)]
    \item the addition of $\mathcal{L}_\text{PDE}$ significantly reduces residuals and in turn improves physical consistency, 
    \item target-free refinement consistently leads to lowest residuals, and
    \item additional gains can be obtained by extending the coverage of the parameter space by augmenting \textbf{D1} with \textbf{D2}.
\end{enumerate*}
In \cref{fig:residual_dist_fine_tuning} we show the distribution of residuals for the different conservation terms to highlight the significant improvement for target-free refinement.
In addition we show in \cref{tab:pred_vs_derived_after_ref} in \cref{app:thermodynamic_consistency} that the model after refinement is more thermodynamically consistent in the sense that derived quantities align better with the ground-truth steady state.
Moreover, in \cref{fig:pointwise_residual_per_term} we visualize the improvement on the different residual terms compared to the base model which highlights that our physics-aware model refinement significantly reduces residuals especially in regions around the shockwaves, indicated by the highlighted blue region.

Finally, we investigate the effect of refinement on the predicted flowfields.
\cref{tab:rel_impr_flowfields_finetune} reports the relative improvement in field-level L2 error for the different refinement variants.
Generally we observe minor fluctuations ($<1\%$) on the derived quantities ($h$, $p_t$, $k$, $T$, $Ma$).
The difference is more pronounced for the primitive fields $p$, $\rho$, and $\mathbf{u}$, where we observe a slight degradation of field errors up to $\sim6\%$ for target-free refinement.
However, this degradation is less pronounced on the OOD set ($\sim1\%$) and comparable to refinement using $\mathcal{L}_\text{data}$.
Our interpretation is that residual-based refinement substantially improves local conservation properties but yields only minor changes in aggregate field accuracy, suggesting that the quality of the pre-trained prior remains the dominant factor in prediction accuracy of the field values.

\begin{table}[tb!]
  \centering
  \caption{Mean per-sample RMS of the conservation-law residuals for mass $r_\rho$, $x$-momentum $r_{\rho u}$, $y$-momentum $r_{\rho v}$, and energy $r_E$, each normalized by its characteristic flux scale, for different models fine-tuned on different loss terms. Lower is more physically consistent. Averaged over three random seeds. $\ast$ denotes extension of training data with additional target-free cases while matching the number of update steps.}
  \vspace{.5em}
  \label{tab:rel_err_residuals_finetune}
  \setlength{\tabcolsep}{6pt}
  \resizebox{\textwidth}{!}{
  \begin{tabular}{llcccc}
  \toprule
  \textbf{Split} & \textbf{Model} & $R_\rho$ & $R_{\rho u}$ & $R_{\rho v}$ & $R_E$ \\
  \midrule
  \multirow{5}{*}{Val}
  & Base                               & $0.2753_{\pm 0.0002}$ & $0.2590_{\pm 0.0001}$ & $0.7055_{\pm 0.0023}$ & $0.3097_{\pm 0.0003}$ \\
  & + Data                             & $0.2721_{\pm 0.0001}$ & $0.2557_{\pm 0.0001}$ & $0.7001_{\pm 0.0000}$ & $0.3060_{\pm 0.0001}$ \\
  & + Data + Residual                  & $0.1271_{\pm 0.0004}$ & $0.1131_{\pm 0.0005}$ & $0.5154_{\pm 0.0004}$ & $0.1363_{\pm 0.0006}$ \\
  \cmidrule(lr){2-6}
  & + Divergence + Residual            & $0.0825_{\pm 0.0000}$ & $0.0734_{\pm 0.0001}$ & $\mathbf{0.3714}_{\pm 0.0000}$ & $0.0856_{\pm 0.0000}$ \\
  & $\text{+ Divergence + Residual}^\ast$ & $\mathbf{0.0806}_{\pm 0.0001}$ & $\mathbf{0.0715}_{\pm 0.0000}$ & $0.3732_{\pm 0.0000}$ & $\mathbf{0.0839}_{\pm 0.0000}$ \\
  \midrule
  \multirow{5}{*}{Test}
  & Base                               & $0.2729_{\pm 0.0001}$ & $0.2573_{\pm 0.0003}$ & $0.6961_{\pm 0.0024}$ & $0.3070_{\pm 0.0010}$ \\
  & + Data                             & $0.2693_{\pm 0.0001}$ & $0.2537_{\pm 0.0001}$ & $0.6910_{\pm 0.0000}$ & $0.3027_{\pm 0.0001}$ \\
  & + Data + Residual                  & $0.1251_{\pm 0.0004}$ & $0.1116_{\pm 0.0005}$ & $0.5093_{\pm 0.0004}$ & $0.1338_{\pm 0.0006}$ \\
  \cmidrule(lr){2-6}
  & + Divergence + Residual            & $0.0810_{\pm 0.0000}$ & $0.0723_{\pm 0.0000}$ & $\mathbf{0.3667}_{\pm 0.0000}$ & $0.0840_{\pm 0.0000}$ \\
  & $\text{+ Divergence + Residual}^\ast$ & $\mathbf{0.0792}_{\pm 0.0001}$ & $\mathbf{0.0705}_{\pm 0.0000}$ & $0.3689_{\pm 0.0000}$ & $\mathbf{0.0824}_{\pm 0.0000}$ \\
  \midrule
  \multirow{5}{*}{OOD}
  & Base                               & $0.3126_{\pm 0.0033}$ & $0.2886_{\pm 0.0026}$ & $0.8219_{\pm 0.0068}$ & $0.3517_{\pm 0.0026}$ \\
  & + Data                             & $0.3120_{\pm 0.0000}$ & $0.2873_{\pm 0.0000}$ & $0.8244_{\pm 0.0005}$ & $0.3506_{\pm 0.0000}$ \\
  & + Data + Residual                  & $0.1601_{\pm 0.0005}$ & $0.1404_{\pm 0.0005}$ & $0.5940_{\pm 0.0005}$ & $0.1716_{\pm 0.0006}$ \\
  \cmidrule(lr){2-6}
  & + Divergence + Residual            & $0.1061_{\pm 0.0000}$ & $0.0932_{\pm 0.0001}$ & $\mathbf{0.4234}_{\pm 0.0001}$ & $0.1105_{\pm 0.0000}$ \\
  & $\text{+ Divergence + Residual}^\ast$ & $\mathbf{0.1042}_{\pm 0.0001}$ & $\mathbf{0.0915}_{\pm 0.0000}$ & $0.4249_{\pm 0.0000}$ & $\mathbf{0.1090}_{\pm 0.0000}$ \\
  \bottomrule
  \end{tabular}}
\end{table}

\begin{table}[tb!]
  \centering
  \caption{Relative change (\%) over the Base model in the relative L2 error of static pressure $p$, density $\rho$, velocity $\bm{u}$, enthalpy $h$, total pressure $p_t$, kinetic energy $k$, temperature $T$, and Mach number $Ma$, for the fine-tuned variants (positive $=$ lower error than Base). Computed from the mean errors over three random seeds. $\ast$ denotes extension of training data with additional target-free cases while matching the number of update steps.}
  \vspace{.5em}
  \label{tab:rel_impr_flowfields_finetune}
  \setlength{\tabcolsep}{6pt}
  \resizebox{\textwidth}{!}{
  \begin{tabular}{llcccccccc}
  \toprule
  \textbf{Split} & \textbf{Model} & $p$ & $\rho$ & $\bm{u}$ & $h$ & $p_t$ & $k$ & $T$ & $Ma$ \\
  \midrule
  \multirow{4}{*}{Val}
  & + Data                             & $+0.75$ & $+0.52$ & $+0.49$ & $-0.49$ & $-0.94$ & $+0.20$ & $-0.49$ & $-0.83$ \\
  & + Data + Residual                  & $+1.95$ & $+0.96$ & $-0.07$ & $-0.17$ & $-1.04$ & $+0.65$ & $-0.17$ & $-0.59$ \\
  \cmidrule(lr){2-10}
  & + Divergence + Residual            & $-3.68$ & $-5.78$ & $-5.95$ & $-0.59$ & $-1.89$ & $-0.53$ & $-0.59$ & $-0.89$ \\
  & $\text{+ Divergence + Residual}^\ast$ & $-3.54$ & $-5.77$ & $-5.98$ & $-0.58$ & $-2.13$ & $-0.45$ & $-0.58$ & $-0.85$ \\
  \midrule
  \multirow{4}{*}{Test}
  & + Data                             & $+0.92$ & $+0.85$ & $+0.93$ & $-0.02$ & $+0.66$ & $+0.82$ & $-0.02$ & $+0.27$ \\
  & + Data + Residual                  & $+2.13$ & $+1.29$ & $+0.35$ & $+0.30$ & $+0.53$ & $+1.26$ & $+0.30$ & $+0.53$ \\
  \cmidrule(lr){2-10}
  & + Divergence + Residual            & $-3.90$ & $-5.55$ & $-5.61$ & $-0.47$ & $-0.75$ & $-0.15$ & $-0.47$ & $+0.03$ \\
  & $\text{+ Divergence + Residual}^\ast$ & $-3.87$ & $-5.60$ & $-5.64$ & $-0.28$ & $-0.74$ & $-0.03$ & $-0.28$ & $+0.24$ \\
  \midrule
  \multirow{4}{*}{OOD}
  & + Data                             & $+0.29$ & $-0.01$ & $+0.41$ & $-0.75$ & $-0.44$ & $+0.02$ & $-0.75$ & $-0.60$ \\
  & + Data + Residual                  & $+2.27$ & $+1.57$ & $+1.31$ & $-0.14$ & $-0.19$ & $+0.77$ & $-0.14$ & $-0.00$ \\
  \cmidrule(lr){2-10}
  & + Divergence + Residual            & $+0.20$ & $-1.57$ & $-1.53$ & $-0.33$ & $-0.73$ & $+0.10$ & $-0.33$ & $+0.15$ \\
  & $\text{+ Divergence + Residual}^\ast$ & $+0.36$ & $-1.42$ & $-1.49$ & $-0.20$ & $-1.23$ & $+0.27$ & $-0.20$ & $+0.19$ \\
  \bottomrule
  \end{tabular}}
\end{table}

\section{Discussion}
\label{sec:discussion}

\textbf{No absolute hierarchy in model choice.}
We observe different trade-offs between the architectures and training paradigms we investigated.
In the data-scarce regime the ViT's regular-grid inductive bias makes it more
sample-efficient than AB-UPT's point-cloud representation. 
Flow matching, while trailing the deterministic models on accuracy, exhibits a substantially
smaller in-distribution-to-OOD performance gap and the strongest correlation between predictive uncertainty and per-sample error on derived quantities. 
Rather than seeking a single best architecture, practitioners should select the model class that matches the data budget, the downstream decision (point estimate versus uncertainty-aware), and the inference-time computational constraint.

\textbf{Physics-aware refinement improves physical consistency.}
Model refinement based on the physics loss consistently improves residuals, while field errors only exhibit slight variations. 
Since residuals are computed via derivatives and non-linear transformations of the state variables, even small modifications to the predicted fields can produce disproportionately larger changes in the residuals.
Most importantly, target-free refinement, only requires meshes and design parameters and never observes a ground-truth field. 
This recovers the physical consistency benefits of PINN-style residual minimization \citep{karniadakis2021physics} while avoiding their well-known optimization difficulties \citep{wang2022and}, as the physics-aware refinement starts from a trained prior.

\textbf{Implications for design optimization.}
Because the PE is fully differentiable and the geometric and inflow conditioning is explicit, the trained emulator can be inserted directly into gradient-based design loops as in \citep{Bezgin_ETAL_25,paischer2025going}, returning sensitivities of integrated KPIs with
respect to the $15$ geometric parameters and the freestream Mach number in milliseconds. 
The error numbers on the OOD set are comparable to typical engineering tolerances in early-stage design, suggesting that the PE could already be deployed for design-space exploration with full-solver verification reserved for selected candidates. 
However, we note that we have not explored our PE in such a setting yet.

\textbf{Practical recommendations.} Based on our findings, we propose a practical guide for developing physics emulators in hypersonic regimes. 
\begin{enumerate*}[label=(\arabic*)]
    \item \textbf{Establish model requirements.} Determine whether the downstream task demands point estimates or uncertainty quantification, what inference latency is acceptable, and how much training data are available. In data-scarce settings, architectures with strong inductive biases such as the regular-grid ViT are preferable. When data are abundant and accuracy is paramount, more expressive pointwise architectures such as AB-UPT should be favored. If predictive uncertainty is needed, probabilistic formulations such as flow matching offer built-in distributional estimates.
    \item \textbf{Establish scaling behavior.} Train at progressively larger dataset sizes and monitor the validation error. As long as the scaling curve has not saturated and the compute budget permits, generating additional data is the most impactful investment, as global prediction quality is determined by the pre-training stage. 
    \item \textbf{ Apply target-free refinement.} Apply residual-based refinement to improve local physical consistency, ideally on top of a well-converged base model. This step requires only meshes and design parameters, no reference flow fields, making it applicable to new regions of the design space at minimal cost. In our experiments, refinement primarily improves local conservation residuals with limited effect on aggregate field accuracy, suggesting it complements rather than replaces pre-training data.
\end{enumerate*}

\section{Limitations}
\label{sec:limitations}

\textbf{Scope of the present study and outlook.}
All scramjet configurations in this work are two-dimensional.
While two-dimensional configurations omit certain physical effects, they already contain sharp discontinuities and multi-scale flow structures that remain highly challenging for neural emulators due to spectral bias \citep{rahaman_2019_spectralbias,xu_2019_fp}.
Extension to 3D, including turbulence, shock-boundary-layer interaction, and finite-rate chemistry, is a natural next step which we have not explored here.
The pre-training dataset comprises roughly $7{,}000$ simulations, which our model-scaling analysis reveals to be the binding constraint beyond ${\sim}25$M parameters.
A promising direction to overcome this bottleneck is pre-training on cheaper, intentionally under-converged data and recovering solution quality through the target-free refinement stage, further reducing the cost of data generation.

\textbf{The differentiability requirement can be a barrier.}
Our workflow requires the solver to be differentiable and the residual to be evaluated with the same discretization used to produce the training data. 
JAX-Fluids meets both requirements, but the wider CFD ecosystem remains dominated by CPU-based, non-differentiable codes. 
The gap is closing through native GPU rewrites in JAX and PyTorch, and through emerging tooling that lowers the porting cost and we expect target-free refinement to become applicable across an increasing fraction of the engineering simulation stack in the future. 
In the interim, compatible differentiable approximations of legacy solvers may offer a useful bridge.

\section{Conclusions}
\label{sec:conclusions}

We introduce a fully GPU-based workflow for neural physics emulators specifically designed for the stringent requirements of hypersonic flow simulation. 
Our workflow ranges from data generation via JAX-Fluids to pre-training and physics-aware refinement leveraging the differentiable code.

To make regular-grid architectures applicable to the adaptive block-structured octree meshes produced by modern GPU-native solvers, we combine absolute and relative positional encodings in physical space, rendering the model agnostic to block count and ordering.
This enables a direct comparison of regular-grid and pointwise paradigms on the same data, where we uncover various trade-offs.
In addition to spatial representation, we compare deterministic and probabilistic training paradigms and conduct scaling studies with respect to model and data size. 
Probabilistic modeling trades uncertainty estimates for accuracy and generally exhibits a narrower gap between in-distribution and OOD performance.
Furthermore, regular-grid methods outperform pointwise methods in data-scarce regimes.
While pointwise methods yield highest accuracy in data-abundant regimes, they are slowest during inference followed by probabilistic and regular-grid methods.
Therefore, the appropriate choice depends on the data budget, downstream requirements, and inference-time constraints.

Our physics-aware refinement depends only on the mesh and design parameters, hence can be carried out in a fully target-free manner without reference flowfields.
This is significant for two reasons. 
First, new regions of the design space can be reached through meshing alone, at a small fraction of the cost compared to data generation. 
Second, the physics loss encourages the surrogate to produce solutions that better satisfy the discretized conservation laws than the base model, which is particularly valuable in extrapolatory regimes where additional training data are unavailable.

Our results demonstrate that a differentiable solver can serve as a refinement engine for neural PE. 
As differentiable GPU-native solvers mature and pre-training datasets grow, this paradigm could extend the practical reach of neural PE to increasingly complex flow regimes. Our workflow is a step toward a broader role for solvers in CFD: not only as primary simulation tools, but as differentiable engines for training and refining physics emulators.

\section*{Acknowledgement}
\label{sec:acknowledgement}

NAA acknowledges funding through ERC Advanced Grant Project No. 101094463.
DAB, ABB, and NAA gratefully acknowledge the Gauss Centre for Supercomputing e.V. (www.gauss-centre.eu) for funding this project by providing computing time on the GCS Supercomputer JUWELS~\citep{JUWELS} at J\"{u}lich Supercomputing Centre (JSC).

\bibliographystyle{alpha}
\bibliography{Hypersonics}

@book{heiser1994hypersonic,
  title={Hypersonic airbreathing propulsion},
  author={Heiser, William H and Pratt, David T},
  year={1994},
  publisher={Aiaa}
}

@article{Urzay18,
   author = "Urzay, Javier",
   title = "Supersonic Combustion in Air-Breathing Propulsion Systems for Hypersonic Flight", 
   journal= "Annual Review of Fluid Mechanics",
   year = "2018",
   volume = "50",
   number = "Volume 50, 2018",
   pages = "593-627",
   doi = "https://doi.org/10.1146/annurev-fluid-122316-045217",
   publisher = "Annual Reviews",
   issn = "1545-4479",
   type = "Journal Article",
  }

@article{McKee80,
   author = "McKee, Christopher P. and Hollenbach, David J.",
   title = "Interstellar Shock Waves", 
   journal= "Annual Review of Astronomy and Astrophysics",
   year = "1980",
   volume = "18",
   number = "Volume 18, 1980",
   pages = "219-262",
   doi = "https://doi.org/10.1146/annurev.aa.18.090180.001251",
   publisher = "Annual Reviews",
   issn = "1545-4282",
   type = "Journal Article",
  }

@article{JagtapMaoAdamsKarniadakis22,
title = {Physics-informed neural networks for inverse problems in supersonic flows},
journal = {Journal of Computational Physics},
volume = {466},
pages = {111402},
year = {2022},
issn = {0021-9991},
doi = {https://doi.org/10.1016/j.jcp.2022.111402},
author = {Ameya D. Jagtap and Zhiping Mao and Nikolaus Adams and George Em Karniadakis},
}

@misc{Toro09,
author = {Toro,Eleuterio F.},
address = {Berlin Heidelberg New York (N.Y.)},
booktitle = {Riemann solvers and numerical methods for fluid dynamics : a practical introduction},
edition = {3rd ed.},
isbn = {978-3-540-25202-3},
language = {eng},
publisher = {Springer-Verlag},
title = {Riemann solvers and numerical methods for fluid dynamics  [Texte imprimé]  : a practical introduction / Eleuterio F. Toro},
year = {2009},
}

@inproceedings{Wilfong_ETAL_25,
author = {Wilfong, Benjamin and Radhakrishnan, Anand and Le Berre, Henry and Vickers, Daniel and Prathi, Tanush and Tselepidis, Nikolaos and Dorschner, Benedikt and Budiardja, Reuben and Cornille, Brian and Abbott, Stephen and Sch\"{a}fer, Florian and Bryngelson, Spencer},
title = {Simulating many-engine spacecraft: Exceeding 1 quadrillion degrees of freedom via information geometric regularization},
year = {2025},
isbn = {9798400714665},
publisher = {Association for Computing Machinery},
address = {New York, NY, USA},
doi = {10.1145/3712285.3771783},
booktitle = {Proceedings of the International Conference for High Performance Computing, Networking, Storage and Analysis},
pages = {14–24},
numpages = {11},
location = {
},
series = {SC '25}
}

@inproceedings{Rossinelli_ETAL_13,
author = {Rossinelli, Diego and Hejazialhosseini, Babak and Hadjidoukas, Panagiotis and Bekas, Costas and Curioni, Alessandro and Bertsch, Adam and Futral, Scott and Schmidt, Steffen J. and Adams, Nikolaus A. and Koumoutsakos, Petros},
title = {11 PFLOP/s simulations of cloud cavitation collapse},
year = {2013},
isbn = {9781450323789},
publisher = {Association for Computing Machinery},
address = {New York, NY, USA},
doi = {10.1145/2503210.2504565},
booktitle = {Proceedings of the International Conference on High Performance Computing, Networking, Storage and Analysis},
articleno = {3},
numpages = {13},
location = {Denver, Colorado},
series = {SC '13}
}

@article{BruntonNoackKoumoutsakos20,
   author = "Brunton, Steven L. and Noack, Bernd R. and Koumoutsakos, Petros",
   title = "Machine Learning for Fluid Mechanics", 
   journal= "Annual Review of Fluid Mechanics",
   year = "2020",
   volume = "52",
   number = "Volume 52, 2020",
   pages = "477-508",
   doi = "https://doi.org/10.1146/annurev-fluid-010719-060214",
   publisher = "Annual Reviews",
   issn = "1545-4479",
   type = "Journal Article",
  }

@article{Bezgin_ETAL_25,
title = {ML-ILES: End-to-end optimization of data-driven high-order Godunov-type finite-volume schemes for compressible homogeneous isotropic turbulence},
journal = {Journal of Computational Physics},
volume = {522},
pages = {113560},
year = {2025},
issn = {0021-9991},
doi = {https://doi.org/10.1016/j.jcp.2024.113560},
author = {Deniz A. Bezgin and Aaron B. Buhendwa and Steffen J. Schmidt and Nikolaus A. Adams},
}

@article{BezginBuhendwaAdams25,
title = {JAX-Fluids 2.0: Towards HPC for differentiable CFD of compressible two-phase flows},
journal = {Computer Physics Communications},
volume = {308},
pages = {109433},
year = {2025},
issn = {0010-4655},
doi = {https://doi.org/10.1016/j.cpc.2024.109433},
author = {Deniz A. Bezgin and Aaron B. Buhendwa and Nikolaus A. Adams},
}

@article{BezginBuhendwaAdams23,
title = {JAX-Fluids: A fully-differentiable high-order computational fluid dynamics solver for compressible two-phase flows},
journal = {Computer Physics Communications},
volume = {282},
pages = {108527},
year = {2023},
issn = {0010-4655},
doi = {https://doi.org/10.1016/j.cpc.2022.108527},
author = {Deniz A. Bezgin and Aaron B. Buhendwa and Nikolaus A. Adams},
}

@book{LeVeque92,
  title={Numerical Methods for Conservation Laws},
  author={LeVeque, R.J.},
  isbn={9783764327231},
  lccn={lc92003400},
  series={Lectures in Mathematics ETH Z{\"u}rich, Department of Mathematics Research Institute of Mathematics},
  year={1992},
  publisher={Springer Basel AG}
}

@article{alkin2025ab,
  title={AB-UPT: Scaling neural CFD surrogates for high-fidelity automotive aerodynamics simulations via anchored-branched universal physics transformers},
  author={Alkin, Benedikt and Bleeker, Maurits and Kurle, Richard and Kronlachner, Tobias and Sonnleitner, Reinhard and Dorfer, Matthias and Brandstetter, Johannes},
  journal={arXiv preprint arXiv:2502.09692},
  year={2025}
}

@inproceedings{dosovitsky2021vit,
	author = {Alexey Dosovitskiy and Lucas Beyer and Alexander Kolesnikov and Dirk Weissenborn and Xiaohua Zhai and Thomas Unterthiner and Mostafa Dehghani and Matthias Minderer and Georg Heigold and Sylvain Gelly and Jakob Uszkoreit and Neil Houlsby},
	booktitle = {{ICLR}},
	title = {An Image is Worth 16x16 Words: Transformers for Image Recognition at Scale},
	year = {2021}
}

@article{ho2020denoising,
  title={Denoising diffusion probabilistic models},
  author={Ho, Jonathan and Jain, Ajay and Abbeel, Pieter},
  journal={Advances in neural information processing systems},
  volume={33},
  pages={6840--6851},
  year={2020}
}

@ARTICLE{VinuesaBrunton22,
	author = {Vinuesa, Ricardo and Brunton, Steven L.},
	title = {Enhancing computational fluid dynamics with machine learning},
	year = {2022},
	journal = {Nature Computational Science},
	volume = {2},
	number = {6},
	pages = {358 – 366},
	doi = {10.1038/s43588-022-00264-7}
}

@ARTICLE{Azizzadenesheli_ETAL_24,
	author = {Azizzadenesheli, Kamyar and Kovachki, Nikola and Li, Zongyi and Liu-Schiaffini, Miguel and Kossaifi, Jean and Anandkumar, Anima},
	title = {Neural operators for accelerating scientific simulations and design},
	year = {2024},
	journal = {Nature Reviews Physics},
	volume = {6},
	number = {5},
	pages = {320 – 328},
	doi = {10.1038/s42254-024-00712-5}
}

@article{GuptaDuraisamy26,
title = {Computational and physical considerations for the development of machine learning augmented turbulence models},
author = {Niloy Gupta and Karthik Duraisamy},
journal = {International Journal of Heat and Fluid Flow},
volume = {117},
pages = {110089},
year = {2026},
issn = {0142-727X},
doi = {https://doi.org/10.1016/j.ijheatfluidflow.2025.110089}
}

@article{Buhendwa_ETAL_25,
  title = {Data-driven shape inference in three-dimensional steady-state supersonic flows: Optimizing a discrete loss with JAX-Fluids},
  author = {Buhendwa, Aaron B. and Bezgin, Deniz A. and Karnakov, Petr and Adams, Nikolaus A. and Koumoutsakos, Petros},
  journal = {Phys. Rev. Fluids},
  volume = {10},
  issue = {8},
  pages = {084902},
  numpages = {21},
  year = {2025},
  month = {Aug},
  publisher = {American Physical Society},
  doi = {10.1103/9wj9-nmr8}
}

@article{AbaidiAdams25,
title = {Exploring denoising diffusion models for compressible fluid field prediction},
journal = {Computers \& Fluids},
volume = {298},
pages = {106665},
year = {2025},
issn = {0045-7930},
doi = {https://doi.org/10.1016/j.compfluid.2025.106665},
author = {R. Abaidi and N.A. Adams}
}

@article{team2023gemini,
  title={Gemini: a family of highly capable multimodal models},
  author={Team, Gemini and Anil, Rohan and Borgeaud, Sebastian and Alayrac, Jean-Baptiste and Yu, Jiahui and Soricut, Radu and Schalkwyk, Johan and Dai, Andrew M and Hauth, Anja and Millican, Katie and others},
  journal={arXiv preprint arXiv:2312.11805},
  year={2023}
}

@article{vaswani2017attention,
  title={Attention is all you need},
  author={Vaswani, Ashish and Shazeer, Noam and Parmar, Niki and Uszkoreit, Jakob and Jones, Llion and Gomez, Aidan N and Kaiser, {\L}ukasz and Polosukhin, Illia},
  journal={Advances in neural information processing systems},
  volume={30},
  year={2017}
}

@article{achiam2023gpt,
  title={Gpt-4 technical report},
  author={Achiam, Josh and Adler, Steven and Agarwal, Sandhini and Ahmad, Lama and Akkaya, Ilge and Aleman, Florencia Leoni and Almeida, Diogo and Altenschmidt, Janko and Altman, Sam and Anadkat, Shyamal and others},
  journal={arXiv preprint arXiv:2303.08774},
  year={2023}
}

@article{sacks1989design,
  title={Design and Analysis of Computer Experiments},
  author={Sacks, Jerome and Welch, William J. and Mitchell, Toby J. and Wynn, Henry P.},
  journal={Statistical Science},
  volume={4},
  number={4},
  pages={409--423},
  year={1989},
  publisher={Institute of Mathematical Statistics},
  doi={10.1214/ss/1177012413}
}

@article{kennedy2001bayesian,
  title={Bayesian Calibration of Computer Models},
  author={Kennedy, Marc C. and O'Hagan, Anthony},
  journal={Journal of the Royal Statistical Society: Series B (Statistical Methodology)},
  volume={63},
  number={3},
  pages={425--464},
  year={2001},
  publisher={Wiley},
  doi={10.1111/1467-9868.00294}
}

@article{Hu2006,
   author = {X Y Hu and B C Khoo and N A Adams and F L Huang},
   doi = {10.1016/j.jcp.2006.04.001},
   issn = {00219991},
   issue = {2},
   journal = {Journal of Computational Physics},
   pages = {553-578},
   title = {A conservative interface method for compressible flows},
   volume = {219},
   year = {2006},
}

@article{Borges2008,
  author  = {Borges, Rafael and Carmona, Monique and Costa, Bruno and Don, Wai Sun},
  title   = {An improved weighted essentially non-oscillatory scheme for hyperbolic conservation laws},
  journal = {Journal of Computational Physics},
  volume  = {227},
  number  = {6},
  pages   = {3191--3211},
  year    = {2008}
}

@INPROCEEDINGS{liu_2008_isolationforest,
  author={Liu, Fei Tony and Ting, Kai Ming and Zhou, Zhi-Hua},
  booktitle={2008 Eighth IEEE International Conference on Data Mining}, 
  title={Isolation Forest}, 
  year={2008},
  volume={},
  number={},
  pages={413-422},
  doi={10.1109/ICDM.2008.17}}

@article{kochkov2021machine,
  title={Machine learning--accelerated computational fluid dynamics},
  author={Kochkov, Dmitrii and Smith, Jamie A and Alieva, Ayya and Wang, Qing and Brenner, Michael P and Hoyer, Stephan},
  journal={Proceedings of the National Academy of Sciences},
  volume={118},
  number={21},
  pages={e2101784118},
  year={2021},
  publisher={National Academy of Sciences}
}

@article{brenner2019perspective,
  title={Perspective on machine learning for advancing fluid mechanics},
  author={Brenner, Michael P and Eldredge, Jeff D and Freund, Jonathan B},
  journal={Physical Review Fluids},
  volume={4},
  number={10},
  pages={100501},
  year={2019},
  publisher={APS}
}

@article{thuerey2020deep,
  title={Deep learning methods for Reynolds-averaged Navier--Stokes simulations of airfoil flows},
  author={Thuerey, Nils and Wei{\ss}enow, Konstantin and Prantl, Lukas and Hu, Xiangyu},
  journal={AIAA journal},
  volume={58},
  number={1},
  pages={25--36},
  year={2020},
  publisher={American Institute of Aeronautics and Astronautics}
}

@article{ray2018artificial,
  title={An artificial neural network as a troubled-cell indicator},
  author={Ray, Deep and Hesthaven, Jan S},
  journal={Journal of computational physics},
  volume={367},
  pages={166--191},
  year={2018},
  publisher={Elsevier}
}

@article{mao2020physics,
  title={Physics-informed neural networks for high-speed flows},
  author={Mao, Zhiping and Jagtap, Ameya D and Karniadakis, George Em},
  journal={Computer Methods in Applied Mechanics and Engineering},
  volume={360},
  pages={112789},
  year={2020},
  publisher={Elsevier}
}

@inproceedings{lienen2024zero,
  title={From zero to turbulence: Generative modeling for 3d flow simulation},
  author={Lienen, Marten and L{\"u}dke, David and Hansen-Palmus, Jan and G{\"u}nnemann, Stephan},
  booktitle={International Conference on Learning Representations},
  volume={2024},
  pages={5203--5220},
  year={2024}
}

@article{alkin2024universal,
  title={Universal physics transformers: A framework for efficiently scaling neural operators},
  author={Alkin, Benedikt and F{\"u}rst, Andreas and Schmid, Simon and Gruber, Lukas and Holzleitner, Markus and Brandstetter, Johannes},
  journal={Advances in Neural Information Processing Systems},
  volume={37},
  pages={25152--25194},
  year={2024}
}

@article{li2020fourier,
  title={Fourier neural operator for parametric partial differential equations},
  author={Li, Zongyi and Kovachki, Nikola and Azizzadenesheli, Kamyar and Liu, Burigede and Bhattacharya, Kaushik and Stuart, Andrew and Anandkumar, Anima},
  journal={arXiv preprint arXiv:2010.08895},
  year={2020}
}

@article{lu2021learning,
  title={Learning nonlinear operators via DeepONet based on the universal approximation theorem of operators},
  author={Lu, Lu and Jin, Pengzhan and Pang, Guofei and Zhang, Zhongqiang and Karniadakis, George Em},
  journal={Nature machine intelligence},
  volume={3},
  number={3},
  pages={218--229},
  year={2021},
  publisher={Nature Publishing Group UK London}
}

@article{pfaff2020learning,
  title={Learning mesh-based simulation with graph networks},
  author={Pfaff, Tobias and Fortunato, Meire and Sanchez-Gonzalez, Alvaro and Battaglia, Peter W},
  journal={arXiv preprint arXiv:2010.03409},
  year={2020}
}

@article{gao2024generative,
  title={Generative learning for forecasting the dynamics of high-dimensional complex systems},
  author={Gao, Han and Kaltenbach, Sebastian and Koumoutsakos, Petros},
  journal={Nature Communications},
  volume={15},
  number={1},
  pages={8904},
  year={2024},
  publisher={Nature Publishing Group UK London}
}

@article{molinaro2024generative,
  title={Generative ai for fast and accurate statistical computation of fluids},
  author={Molinaro, Roberto and Lanthaler, Samuel and Raoni{\'c}, Bogdan and Rohner, Tobias and Armegioiu, Victor and Simonis, Stephan and Grund, Dana and Ramic, Yannick and Wan, Zhong Yi and Sha, Fei and others},
  journal={arXiv preprint arXiv:2409.18359},
  year={2024}
}

@article{karniadakis2021physics,
  title={Physics-informed machine learning},
  author={Karniadakis, George Em and Kevrekidis, Ioannis G and Lu, Lu and Perdikaris, Paris and Wang, Sifan and Yang, Liu},
  journal={Nature Reviews Physics},
  volume={3},
  number={6},
  pages={422--440},
  year={2021},
  publisher={Nature Publishing Group UK London}
}

@article{liu2024uncertainty,
  title={Uncertainty-aware surrogate models for airfoil flow simulations with denoising diffusion probabilistic models},
  author={Liu, Qiang and Thuerey, Nils},
  journal={AIAA Journal},
  volume={62},
  number={8},
  pages={2912--2933},
  year={2024},
  publisher={American Institute of Aeronautics and Astronautics}
}

@article{fischer2025optimal,
  title={Optimal Lattice Boltzmann Closures through Multi-Agent Reinforcement Learning},
  author={Fischer, Paul and Kaltenbach, Sebastian and Litvinov, Sergey and Succi, Sauro and Koumoutsakos, Petros},
  journal={arXiv preprint arXiv:2504.14422},
  year={2025}
}

@article{novati2021automating,
  title={Automating turbulence modelling by multi-agent reinforcement learning},
  author={Novati, Guido and De Laroussilhe, Hugues Lascombes and Koumoutsakos, Petros},
  journal={Nature Machine Intelligence},
  volume={3},
  number={1},
  pages={87--96},
  year={2021},
  publisher={Nature Publishing Group UK London}
}

@techreport{slotnick2014cfd,
  title={CFD vision 2030 study: a path to revolutionary computational aerosciences},
  author={Slotnick, Jeffrey P and Khodadoust, Abdollah and Alonso, Juan and Darmofal, David and Gropp, William and Lurie, Elizabeth and Mavriplis, Dimitri J},
  year={2014}
}

@article{su2024roformer,
  title={Roformer: Enhanced transformer with rotary position embedding},
  author={Su, Jianlin and Ahmed, Murtadha and Lu, Yu and Pan, Shengfeng and Bo, Wen and Liu, Yunfeng},
  journal={Neurocomputing},
  volume={568},
  pages={127063},
  year={2024},
  publisher={Elsevier}
}

@article{hendrycks2016gaussian,
  title={Gaussian error linear units (gelus)},
  author={Hendrycks, Dan and Gimpel, Kevin},
  journal={arXiv preprint arXiv:1606.08415},
  year={2016}
}

@inproceedings{peebles2023scalable,
  title={Scalable diffusion models with transformers},
  author={Peebles, William and Xie, Saining},
  booktitle={Proceedings of the IEEE/CVF international conference on computer vision},
  pages={4195--4205},
  year={2023}
}

@article{zhang2019root,
  title={Root mean square layer normalization},
  author={Zhang, Biao and Sennrich, Rico},
  journal={Advances in neural information processing systems},
  volume={32},
  year={2019}
}

@article{shazeer2020glu,
  title={Glu variants improve transformer},
  author={Shazeer, Noam},
  journal={arXiv preprint arXiv:2002.05202},
  year={2020}
}

@article{elfwing2018sigmoid,
  title={Sigmoid-weighted linear units for neural network function approximation in reinforcement learning},
  author={Elfwing, Stefan and Uchibe, Eiji and Doya, Kenji},
  journal={Neural networks},
  volume={107},
  pages={3--11},
  year={2018},
  publisher={Elsevier}
}

@article{chen2023symbolic,
  title={Symbolic discovery of optimization algorithms},
  author={Chen, Xiangning and Liang, Chen and Huang, Da and Real, Esteban and Wang, Kaiyuan and Pham, Hieu and Dong, Xuanyi and Luong, Thang and Hsieh, Cho-Jui and Lu, Yifeng and others},
  journal={Advances in neural information processing systems},
  volume={36},
  pages={49205--49233},
  year={2023}
}

@article{li2025back,
  title={Back to basics: Let denoising generative models denoise},
  author={Li, Tianhong and He, Kaiming},
  journal={arXiv preprint arXiv:2511.13720},
  year={2025}
}

@article{lipman2022flow,
  title={Flow matching for generative modeling},
  author={Lipman, Yaron and Chen, Ricky TQ and Ben-Hamu, Heli and Nickel, Maximilian and Le, Matt},
  journal={arXiv preprint arXiv:2210.02747},
  year={2022}
}

@inproceedings{rahaman_2019_spectralbias,
  author       = {Nasim Rahaman and
                  Aristide Baratin and
                  Devansh Arpit and
                  Felix Draxler and
                  Min Lin and
                  Fred A. Hamprecht and
                  Yoshua Bengio and
                  Aaron C. Courville},
  editor       = {Kamalika Chaudhuri and
                  Ruslan Salakhutdinov},
  title        = {On the Spectral Bias of Neural Networks},
  booktitle    = {Proceedings of the 36th International Conference on Machine Learning,
                  {ICML} 2019, 9-15 June 2019, Long Beach, California, {USA}},
  series       = {Proceedings of Machine Learning Research},
  pages        = {5301--5310},
  publisher    = {{PMLR}},
  year         = {2019},
  bibsource    = {dblp computer science bibliography, https://dblp.org}
}

@article{xu_2019_fp,
  author       = {Zhi{-}Qin John Xu and
                  Yaoyu Zhang and
                  Tao Luo and
                  Yanyang Xiao and
                  Zheng Ma},
  title        = {Frequency Principle: Fourier Analysis Sheds Light on Deep Neural Networks},
  journal      = {CoRR},
  volume       = {abs/1901.06523},
  year         = {2019},
  eprinttype   = {arXiv},
  eprint       = {1901.06523},
  bibsource    = {dblp computer science bibliography, https://dblp.org}
}

@article{paischer2025going,
  title={Going with the Speed of Sound: Pushing Neural Surrogates into Highly-turbulent Transonic Regimes},
  author={Paischer, Fabian and Cotteleer, Leo and Dreze, Yann and Kurle, Richard and Rubini, Dylan and Bleeker, Maurits and Kronlachner, Tobias and Brandstetter, Johannes},
  journal={arXiv preprint arXiv:2511.21474},
  year={2025}
}

@article{wang2022and,
  title={When and why PINNs fail to train: A neural tangent kernel perspective},
  author={Wang, Sifan and Yu, Xinling and Perdikaris, Paris},
  journal={Journal of Computational Physics},
  volume={449},
  pages={110768},
  year={2022},
  publisher={Elsevier}
}

@article{jaber_gpunative_2026,
author = {Jaber, Khodr and Essel, Ebenezer and Sullivan, Pierre},
year = {2026},
month = {04},
pages = {110155},
title = {GPU-native Embedding of Complex Geometries in Adaptive Octree Grids Applied to the Lattice Boltzmann Method},
volume = {324},
journal = {Computer Physics Communications},
doi = {10.1016/j.cpc.2026.110155}
}

@misc{carreon2025gpubasedcompressiblecombustionsolver,
      title={A GPU-based Compressible Combustion Solver for Applications Exhibiting Disparate Space and Time Scales}, 
      author={Anthony Carreon and Jagmohan Singh and Shivank Sharma and Shuzhi Zhang and Venkat Raman},
      year={2025},
      eprint={2510.23993},
      archivePrefix={arXiv},
      primaryClass={cs.DC},
      url={https://arxiv.org/abs/2510.23993}, 
}

@article{JUWELS,
author = {{Juelich Supercomputing Centre}},
title = {{JUWELS Cluster and Booster: Exascale Pathfinder with Modular Supercomputing Architecture at Juelich Supercomputing Centre}},
journal = {Journal of large-scale research facilities},
number = {A183},
volume = {7},
doi = {10.17815/jlsrf-7-183},
url = {http://dx.doi.org/10.17815/jlsrf-7-183},
year = {2021}
}

\appendix
\section{Physical and Numerical Model}
\label{app:phys_model}

The fluid state at position $\mathbf{x} = [x,y]^T$ and time $t$ is described either by the vector of primitive variables $\mathbf{W} = [\rho, u, v, p]^T$ or by the vector of conservative variables $\mathbf{U} = [\rho, \rho u, \rho v, E]^T$.
Here, $\rho$ denotes the density, $u$ and $v$ are the velocity components in the $x$- and $y$-directions, respectively, and $p$ is the pressure.
The velocity vector is denoted by $\mathbf{u} = [u,v]^T$.
The total energy per unit volume is given by
\begin{align}
    E = \rho e + \frac{1}{2}\rho \mathbf{u} \cdot \mathbf{u},
\end{align}
where $e$ is the specific internal energy.

The present study considers inviscid compressible flows governed by the two-dimensional Euler equations,
\begin{align}
    \frac{\partial \mathbf{U}}{\partial t}
    +
    \frac{\partial \mathbf{F}(\mathbf{U})}{\partial x}
    +
    \frac{\partial \mathbf{G}(\mathbf{U})}{\partial y}
    = 0,
    \label{eq:euler}
\end{align}
where $\mathbf{F}$ and $\mathbf{G}$ are the convective flux vectors in $x$- and $y$-directions, respectively,
\begin{align}
    \mathbf{F}(\mathbf{U}) =
    \begin{bmatrix}
        \rho u \\
        \rho u^2 + p \\
        \rho u v \\
        u(E+p)
    \end{bmatrix},
    \qquad
    \mathbf{G}(\mathbf{U}) =
    \begin{bmatrix}
        \rho v \\
        \rho u v \\
        \rho v^2 + p \\
        v(E+p)
    \end{bmatrix}.
    \label{eq:convective_fluxes}
\end{align}
The system is closed using the ideal-gas equation of state,
\begin{align}
    p = (\gamma - 1)\rho e,
    \qquad
    c = \sqrt{\gamma \frac{p}{\rho}},
    \label{eq:ideal_gas_eos}
\end{align}
where $\gamma$ is the ratio of specific heats and $c$ is the speed of sound.
In this work, we use $\gamma = 1.4$.
Equivalently, the pressure can be recovered from the conservative variables as
\begin{align}
    p = (\gamma - 1)
    \left(
        E - \frac{1}{2}\rho(u^2 + v^2)
    \right).
    \label{eq:pressure_from_conservative}
\end{align}

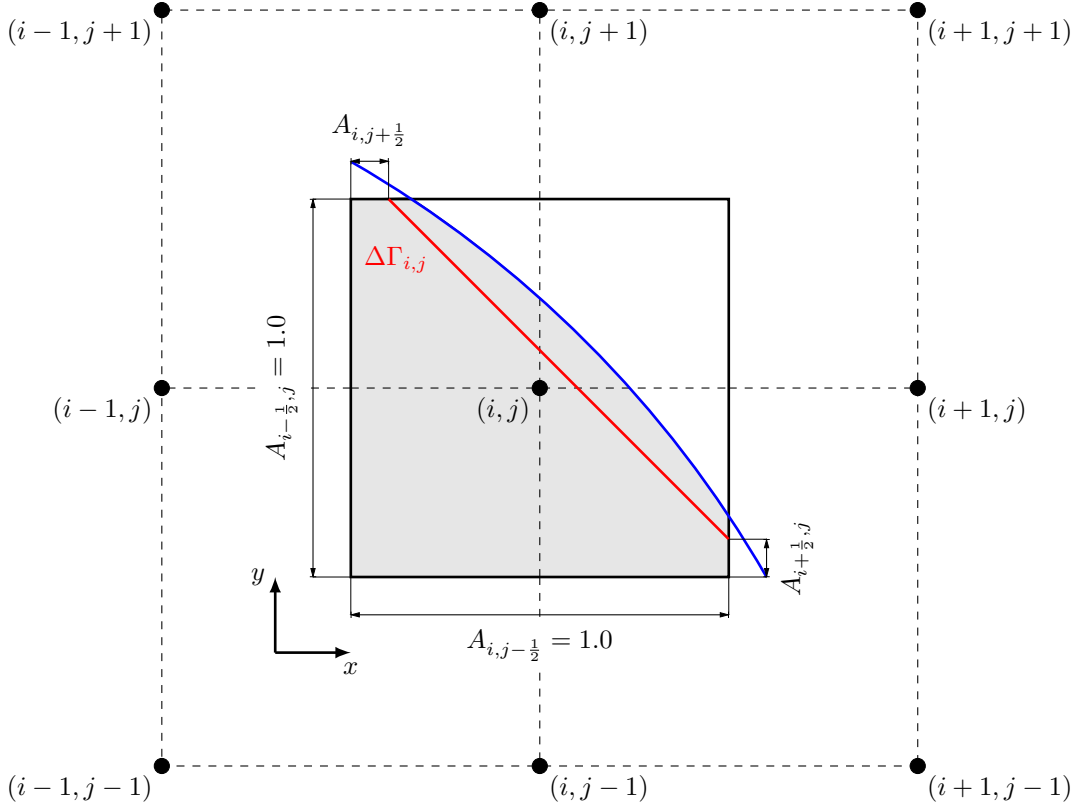
\begin{figure}
    \centering
    \begin{tikzpicture}

    \definecolor{color0}{RGB}{0,101,189}
    \definecolor{color1}{RGB}{227,114,34}
    \definecolor{color2}{RGB}{162,173,0}

    \coordinate (NULL) at (0,0);
    \coordinate (B) at (5,5);
    \coordinate (C) at ($0.5*(B)$);
    \coordinate (D1) at ($0.125*(B)$);
    \coordinate (D2) at ($0.71*(B)$);

    \coordinate (D11) at ($0.1*(B)$);

    \coordinate (L) at ($1.1*(B|-NULL)+0.96*(B-|NULL)$);

    \filldraw[fill=black!10!white, line width=0.1pt] (NULL) -- (B|-NULL) -- ($(B|-NULL) + 0.16*(B-|NULL)$) -- ($(B|-NULL) + 0.16*(B-|NULL)$) arc (33.5:56.4:15) -- (B-|NULL) -- (NULL);

    \draw[line width=1pt] (0,0) rectangle (B);
    \draw[dashed] (C |- NULL) -- (C |- B);
    \draw[dashed] (C -| NULL) -- (C -| B);

    \draw[dashed] ($(C)+(C-|NULL)$) -- ($(C)+(B-|NULL)$);
    \draw[dashed] ($(C)+(C|-NULL)$) -- ($(C)+(B|-NULL)$);
    \draw[dashed] ($(C)-(C-|NULL)$) -- ($(C)-(B-|NULL)$);
    \draw[dashed] ($(C)-(C|-NULL)$) -- ($(C)-(B|-NULL)$);
    \draw[dashed] ($3*(C|-NULL)-(NULL|-C)$) -- ($-1*(C)$) -- ($3*(C-|NULL)-(NULL-|C)$) -- ($3*(C)$) -- cycle;

    \draw[fill=black, draw=black] (C) circle (0.1) node[below left] {$(i,j)$};
    \draw[fill=black, draw=black] ($(C)+(B-|NULL)$) circle (0.1) node[below right] {$(i,j+1)$};
    \draw[fill=black, draw=black] ($(C)+(B|-NULL)$) circle (0.1) node[below right] {$(i+1,j)$};
    \draw[fill=black, draw=black] ($(C)-(B|-NULL)$) circle (0.1) node[below left] {$(i-1,j)$};
    \draw[fill=black, draw=black] ($(C)-(B-|NULL)$) circle (0.1) node[below right] {$(i,j-1)$};
    \draw[fill=black, draw=black] ($-1*(C)$) circle (0.1) node[below left] {$(i-1,j-1)$};
    \draw[fill=black, draw=black] ($3*(C)$) circle (0.1) node[below right] {$(i+1,j+1)$};
    \draw[fill=black, draw=black] ($3*(C-|NULL)-(NULL-|C)$) circle (0.1) node[below left] {$(i-1,j+1)$};
    \draw[fill=black, draw=black] ($3*(C|-NULL)-(NULL|-C)$) circle (0.1) node[below right] {$(i+1,j-1)$};

    \draw[blue, line width=1pt] ($1.1*(B|-NULL)$) arc (30:60:15); \label{tikz:interface}

    \draw[red, line width=1pt] (B -| D11) -- (D11 -| B) node[at start, below, yshift=-0.5cm, xshift=0.1cm] {$\Delta\Gamma_{i,j}$}; \label{tikz:interface_reconstruction}
    
    \coordinate (DIST) at (0.5,0.5);
    \dimline[extension start length=1cm, extension end length=1cm,extension style={black}, label style={above=0.5ex}] {(-0.5,0)}{($(NULL|-B) - (0.5,0)$)}{$A_{i-\frac{1}{2},j}=1.0$};
    \dimline[extension start length=-1cm, extension end length=-1cm,extension style={black}, label style={below=0.5ex}] {(0.0,-0.5)}{($(NULL-|B) - (0.0,0.5)$)}{$A_{i,j-\frac{1}{2}}=1.0$};
    \dimline[extension start length=0.5cm, extension end length=0.5cm,extension style={black}, label style={above=0.5ex}] {($(NULL|-B) + (DIST -| NULL)$)}{($(NULL|-B) + (DIST -| NULL) + (D11|-NULL)$)}{$A_{i,j+\frac{1}{2}}$};
    \dimline[extension start length=-0.5cm, extension end length=-0.5cm,extension style={black}, label style={below=0.5ex}] {($(NULL-|B) + (DIST |- NULL)$)}{($(NULL-|B) + (DIST|- NULL) + (D11-|NULL)$)}{$A_{i+\frac{1}{2},j}$};

    \draw[->, line width=1pt] (-1,-1) -- (-1,0) node[left] {$y$};
    \draw[->, line width=1pt] (-1,-1) -- (0,-1) node[below] {$x$};

\end{tikzpicture}
    \caption{Schematic of a cut cell.}
    \label{fig:cutcell}
\end{figure}

The compressible Euler equations are discretized on a Cartesian grid using a high-order Godunov-type finite-volume formulation~\cite{Toro09}.
The computational domain is divided into rectangular control volumes with uniform cell sizes $\Delta x$ and $\Delta y$.
Cell centers are indexed by $(i,j)$, and $\mathbf{U}_{i,j}$ denotes the cell averaged conservative state.
Numerical fluxes through the vertical and horizontal cell faces are denoted by $\mathbf{F}_{i\pm\frac{1}{2},j}$ and $\mathbf{G}_{i,j\pm\frac{1}{2}}$, respectively.

The numerical fluxes are computed using a two-step high-order Godunov procedure.
First, left and right states are reconstructed at each cell face from the neighboring cell averages using a shock-capturing reconstruction scheme.
In this work, we employ the fifth-order WENO-Z reconstruction~\cite{Borges2008}.
Second, the reconstructed states are used as input to an approximate Riemann solver.
Here, we use the HLLC Riemann solver~\cite{Toro09} to compute the numerical fluxes across the cell faces.

Immersed solid boundaries are represented using a conservative sharp-interface cut-cell method~\cite{Hu2006}.
The solid-fluid interface is described implicitly by a level-set function $\phi(\mathbf{x})$, where $\phi$ satisfies the signed-distance property $\|\nabla \phi\| = 1$.
The interface location is given by the zero level set
\begin{align}
    \Gamma = \left\{ \mathbf{x} \mid \phi(\mathbf{x}) = 0 \right\}.
\end{align}
Cells intersected by the interface are referred to as \textit{cut cells}, while cells that are entirely filled with fluid are referred to as \textit{full cells}.
A schematic illustration of a cut cell, including the fluid volume fraction, face apertures, and interface segment, is shown in \cref{fig:cutcell}.

For each cell, we define the fluid volume fraction $\alpha_{i,j} \in [0,1]$, which denotes the fraction of the cell area occupied by fluid.
We also define the face apertures $A_{i\pm\frac{1}{2},j}$ and $A_{i,j\pm\frac{1}{2}}$, which denote the fractions of the corresponding cell faces open to the fluid phase.
For full fluid cells, all apertures and the volume fraction are equal to one.
The standard finite-volume discretization is therefore recovered as a special case of the cut-cell formulation.

The semi-discrete conservative update for both full cells and cut cells is written as
\begin{align}
    \frac{d}{dt}\left(\alpha_{i,j}\mathbf{U}_{i,j}\right)
    =
    &\frac{1}{\Delta x}
    \left(
        A_{i-\frac{1}{2},j}\mathbf{F}_{i-\frac{1}{2},j}
        -
        A_{i+\frac{1}{2},j}\mathbf{F}_{i+\frac{1}{2},j}
    \right)
    \nonumber\\
    &+
    \frac{1}{\Delta y}
    \left(
        A_{i,j-\frac{1}{2}}\mathbf{G}_{i,j-\frac{1}{2}}
        -
        A_{i,j+\frac{1}{2}}\mathbf{G}_{i,j+\frac{1}{2}}
    \right)
    +
    \frac{1}{\Delta x \Delta y}\mathbf{X}_{i,j}.
    \label{eq:FVD_levelset}
\end{align}
Here, the cell face fluxes are weighted by the corresponding apertures, and $\mathbf{X}_{i,j}$ denotes the interface flux contribution inside cell $(i,j)$.
For full cells, no solid-fluid interface is present and therefore $\mathbf{X}_{i,j} = \mathbf{0}$.
Together with $\alpha_{i,j}=1$ and unit apertures, Eq.~\eqref{eq:FVD_levelset} reduces to the standard semi-discrete finite-volume update on a Cartesian grid.

The interface flux represents the force exerted by the immersed boundary on the fluid.
In two dimensions, it is given by
\begin{align}
    \mathbf{X}_{i,j}
    =
    \begin{bmatrix}
        0 \\
        p_\Gamma \Delta \Gamma_x \\
        p_\Gamma \Delta \Gamma_y \\
        p_\Gamma \Delta \boldsymbol{\Gamma}_{i,j} \cdot \mathbf{v}_\Gamma
    \end{bmatrix},
    \qquad
    \Delta \boldsymbol{\Gamma}_{i,j}
    =
    \begin{bmatrix}
        \Delta \Gamma_x \\
        \Delta \Gamma_y
    \end{bmatrix}
    =
    \begin{bmatrix}
        \left(A_{i+\frac{1}{2},j} - A_{i-\frac{1}{2},j}\right)\Delta y \\
        \left(A_{i,j+\frac{1}{2}} - A_{i,j-\frac{1}{2}}\right)\Delta x
    \end{bmatrix}.
    \label{eq:interface_flux}
\end{align}
Here, $p_\Gamma$ is the pressure at the interface, $\mathbf{v}_\Gamma$ is the interface velocity, and $\Delta \boldsymbol{\Gamma}_{i,j}$ is the oriented interface length vector associated with the cut cell.
The components of this vector correspond to the projections of the interface segment onto the coordinate directions and are obtained from the differences of opposite face apertures.
The interface pressure is approximated by the cell-center pressure of the corresponding cut cell.
Since only static solid bodies are considered in this work, the interface velocity is zero, $\mathbf{v}_\Gamma = \mathbf{0}$, and the energy contribution of the interface flux vanishes.
The interface flux therefore contributes only to the momentum equations through the pressure force exerted by the solid boundary.

The volume fraction $\alpha_{i,j}$, the face apertures $A_{i\pm\frac{1}{2},j}$ and $A_{i,j\pm\frac{1}{2}}$, and the oriented interface length vector $\Delta \boldsymbol{\Gamma}_{i,j}$ are computed from the level-set representation of the geometry using a marching-squares reconstruction.
This yields a conservative cut-cell discretization that preserves the structured Cartesian layout while allowing immersed solid boundaries to be represented sharply.

\section{Data Generation}
\label{app:data_gen}

\begin{figure}
    \centering
    \includegraphics[width=\linewidth]{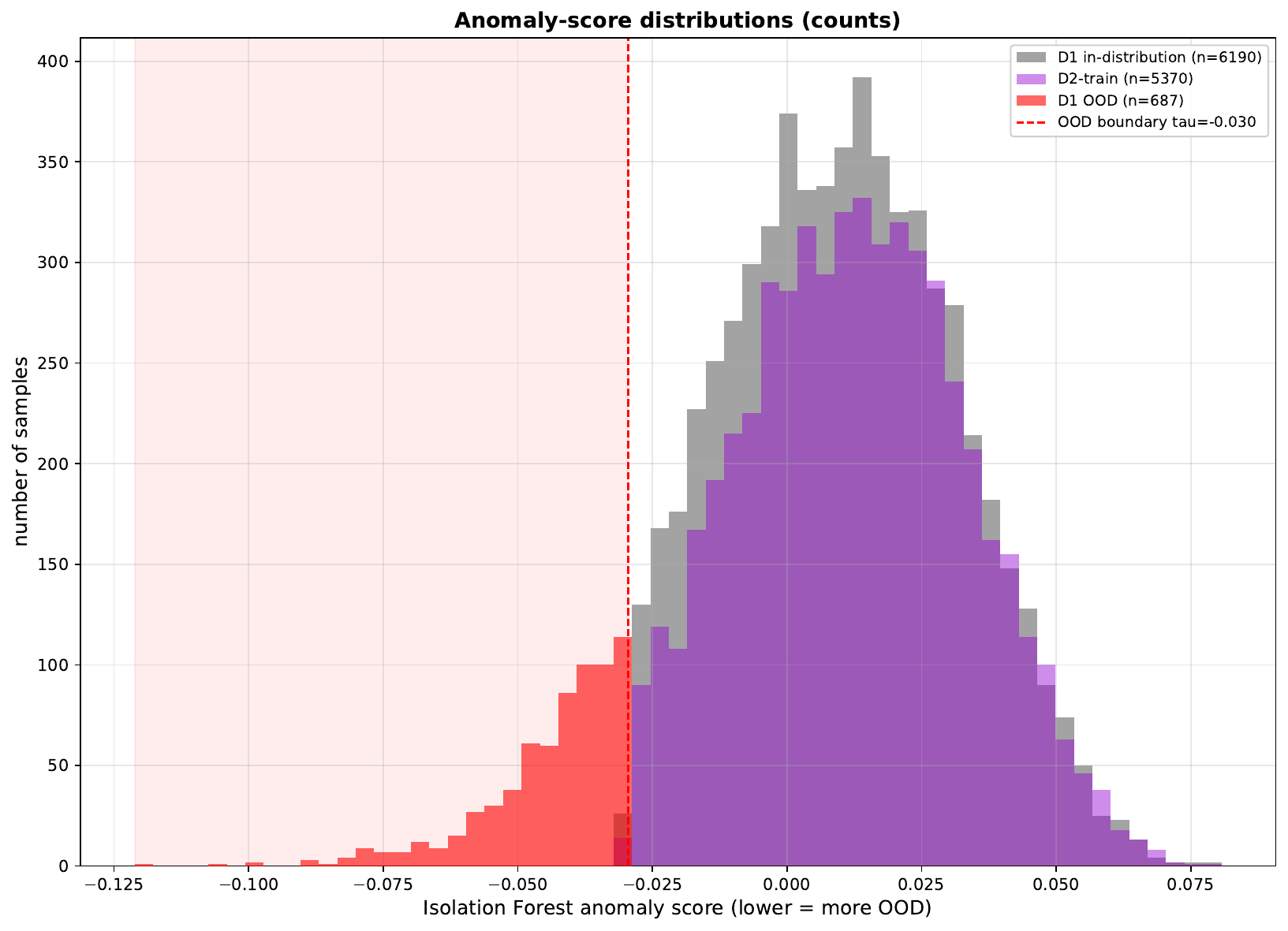}
    \caption{\textbf{Isolation-Forest diagnostic for the out-of-distribution split.} 
    We report the distribution of anomaly scores obtained via Isolation Forest fitted on the geometry and inflow parameters.
    Samples at the lower end of the distribution are selected to be OOD.
    Importantly there is no overlap in the parameter vectors between $D2$ and the OOD test set, so no information is leaked to the OOD set.
    The remaining sets are randomly selected from the remaining samples.
    }
    \label{fig:ood_splits}
\end{figure}

To split the dataset into in-distribution and out-of-distribution subsets without imposing an arbitrary cutoff on any single design variable, we use an unsupervised Isolation Forest \citep{liu_2008_isolationforest} over the $d$-dimensional vector of varying scramjet design parameters.
The forest comprises 100 isolation trees fit with a fixed random seed.
The Isolation Forest builds 100 random binary trees that recursively split the design space, and assigns each sample a score $s_i$ equal to the average number of splits needed to isolate it. 
Samples in sparse regions are isolated in few splits and therefore receive low scores, which we use to flag out-of-distribution candidates.
We designated as out-of-distribution the $k$ samples with the lowest scores, where $k$ is fixed by the requested OOD fraction. 
\cref{fig:ood_splits} shows the distribution of the anomaly scores over all cases and the different datasets ($D1,D2$) to illustrate that the selected OOD split generally corresponds to the tail of the distribution, so edge cases of the generated data.
While there is minimal overlap in the anomaly scores between $D2$ and the selected OOD samples we stress that these do not correspond to exact same parameter settings.
In fact, there is no exact match between any of the cases in the overlapping anomaly scores in terms of geometry and inflow parameters so the parameter space they cover differ.

\section{Implementation Details}
\label{app:impl_dets}

\subsection{AB-UPT}

We instantiate AB-UPT as a volume-only configuration, which discards the surface branch and operates solely on a 3D volumetric anchor set, since the scramjet task requires field predictions inside the flow domain rather than on a surface. 
From every simulation we draw $N_{\mathrm{vol}}=16{,}384$ volume anchor points uniformly at random from the mesh.
Volumetric coordinates are linearly rescaled into $[0, 1000]^2$ before being fed to the network, and we predict all eight target fields, namely static pressure, density, two-component velocity, specific enthalpy, total pressure, kinetic energy, temperature and Mach number.
Each field is standardized per-channel using statistics computed once over the training split.
Each anchor position $\mathbf{x}_i\in\mathbb{R}^2$ is first lifted to a token of width $d=384$ via a continuous sin–cos Fourier embedding followed by a two-layer GELU MLP \citep{hendrycks2016gaussian}.
We additionally apply rotary positional encodings \citep[RoPE]{su2024roformer} inside attention. 
The trunk is a stack of ten self-attention blocks followed by four additional volume-decoder blocks.
Therefore, during training anchors are used as both queries and keys/values.
We select $h=3$ heads, an MLP expansion factor of 4, and truncated-normal initialization ($\sigma=0.02$).
A final linear head decodes each anchor token to the eight-channel field prediction. The model is conditioned on a 16-dimensional design vector (15 geometry parameters and the inflow Mach number) that is injected into every transformer block via the AdaLN-style scale/shift pathway \citep{peebles2023scalable} with a per-condition dimension of 16. 
Training minimizes an unweighted sum of channel-wise MSE losses on the standardized targets, optimized with Lion \citep{chen2023symbolic} at a peak learning rate of $1\times10^{-5}$, weight decay $0.05$, gradient-norm clipping at $1.0$, and a linear warm-up over $5\%$ of training followed by a cosine decay to $1\times10^{-6}$. 
We train for 250 epochs in float16 mixed precision with an effective batch size of one simulation per step, where each step performs attention over $16{,}384$ anchor tokens. We select the final checkpoint by the best average relative $L_2$ loss on the  validation set.

For scaling experiments we vary the depth of the model ranging from $\approx 10M$ parameters to $\approx 100M$ parameters.
Specifically, we vary the depth in terms of transformer blocks in $\{ 1, 10, 25, 50\}$ as it allows higher learning rates for larger models compared to scaling in width.
For each of these variants we keep the same model hidden dimension as $d=384$.

\subsection{ViT}

We compare against a grid-native Vision Transformer \citep[ViT]{dosovitsky2021vit} tailored to the block-structured, obtained by adaptively-refined meshes produced by the JAX-Fluids solver.
Each simulation is exposed to the network as a stack of $N_B$ Cartesian blocks of shape $H\times W$ (in our case $H=W=64$), with per-block isotropic spatial scale and a Boolean fluid mask flagging cells outside the wetted domain. 
Coordinates are min–max rescaled to $[0, 1000]^2$ and per-block scales are min–max rescaled to $[0.2, 1000]$ before being passed to the network.
All eight target fields are channel-wise z-score normalized using the same precomputed statistics as for the AB-UPT model. 
Patchification is performed with a non-overlapping patch size of $p=16$ via 2D average pooling on the coordinate grid.
Blocks with zero fluid fraction are pruned and the remaining blocks are right-padded across the batch so that token sequences are concatenated over all blocks of a sample. 
Each patch centroid is then encoded with a continuous sin–cos Fourier embedding followed by a SiLU MLP \citep{elfwing2018sigmoid} into a token of width $d=384$, and RoPE is applied inside every attention head, derived from the same average-pooled centers. 

The backbone is a stack of $L=8$ pre-norm transformer blocks with $h=3$ heads, RMSNorm ($\varepsilon=10^{-6}$) \citep{zhang2019root}, a SwiGLU FFN \citep{shazeer2020glu} with expansion factor 4, and AdaLN-Zero conditioning on the 16-dimensional design vector.
The design vector is first embedded by a shared SiLU MLP and then drives a per-block linear head that produces six modulation parameters (shift, scale, gate for MSA and MLP) per channel. 
Weights are initialized with Xavier-uniform on all linear layers and the AdaLN modulation projection and the final linear layer are zero-initialized, while the MSA/MLP gate biases are reset to $1.0$ so blocks are fully active at step zero and the network does not collapse to the identity (a failure mode we observed under the standard AdaLN-Zero initialization in this regression setting). 
The output head projects each patch token to $p^2\cdot C_\text{out}$ channels which are unpatchified back to the original cell grid and sliced into the eight field-specific predictions.

The training objective is identical to AB-UPT, an unweighted sum of eight per-field MSE losses on the standardized targets and optimized with Lion at peak learning rate $1\times10^{-4}$, weight decay $0.05$, gradient-norm clipping at $1.0$, and a linear warm-up over $5\%$ of training followed by cosine decay to $1\times10^{-6}$. 
We train for 250 epochs in float16 mixed precision with an effective batch size of one simulation per step and select the final checkpoint on the validation loss.

\subsection{Flow Matching}

The third method replaces the deterministic regression objective with a flow-matching generative model on the same multi-block grid representation. 
We adopt the linear stochastic-interpolant formulation \citep{lipman2022flow,li2025back}.
For a clean sample $x_0\in\mathbb{R}^{B\times N_B\times H\times W\times C}$ we collect all eight target fields (velocity, density, pressure, enthalpy, total pressure, kinetic energy, temperature, Mach number) and Gaussian noise $\varepsilon\sim\mathcal N(0, I)$.
Then we sample a timestep $t\sim\mathcal U[0,1]$ uniformly per sample and form the straight-line interpolant $z_t=t,x_0+(1-t)\varepsilon$.
Given a conditioning vector $c$, the network is parameterized to predict the clean state $x_0^\text{pred}=f_\theta(z_t, t, c)$ rather than the velocity or the noise, and we train under the analytic velocity-matching loss $\mathcal L=\frac{1}{|\Omega|}\sum_{\Omega}\big|(x_0-z_t)/(1-t)-(x_0^\text{pred}-z_t)/(1-t)\big|^2$. 
Training is done only on fluid cells $\Omega$ defined by the AMR fluid mask and with $(1-t)$ clamped from below by $t_\varepsilon=5\times10^{-2}$ for numerical safety near $t=1$.
The corruption is i.i.d. standard Gaussian across cells, blocks and channels. 

The denoiser shares the same backbone as the deterministic ViT baseline ($d=384$, 8 transformer blocks, 3 heads, MLP ratio 4, RMSNorm + SwiGLU + AdaLN-Zero, 2D RoPE on average-pooled patch centres) and differs only in the input and conditioning pipelines. 
The noisy field $z_t$ is patchified by a two-stage bottleneck patch embedding with stride $p=16$ projecting the eight target channels to a PCA-like bottleneck of width 64, followed by a $1\times1$ convolution back to width $d=384$.
The scalar timestep is mapped through a sinusoidal embedding of width 256 and a SiLU MLP into a vector of width $d$.
The 16-dimensional design vector (15 geometry parameters concatenated with the inflow Mach number) is independently embedded by a SiLU MLP of width $d$, and the two are summed to form the AdaLN-Zero modulation token fed into every transformer block.
The final layer predicts $p^2\cdot C_\text{out}$ patch channels which are unpatchified back to the original cell grid. 

We train for 250 epochs in bfloat16 mixed precision with an effective batch size of one simulation per step, using Lion at peak learning rate $3\times10^{-6}$, weight decay $0.05$, gradient-norm clipping at $1.0$, a $5\%$ linear warm-up and a cosine decay to $1\times10^{-6}$.
At inference we draw $x_T\sim\mathcal N(0, I)$ and integrate the learned probability flow ODE $\mathrm dx/\mathrm dt=(x_0^\text{pred}(x_t,t)-x_t)/(1-t)$ from $t=0$ to $t=1$ with explicit Euler for $N-1$ steps and a closed-form last step $x\leftarrow\alpha x+(1-\alpha)x_0^\text{pred}$ with $\alpha=(1-t_\text{next})/(1-t_\text{cur})$ that lands the trajectory exactly on the model's terminal $x_0$ estimate.
In practice we always sweep over a variety of integration steps and always report the one resulting in the best average relative $L_2$ error across all field predictions on the validation set.

\subsection{Physics-aware Refinement}
\label{app:physics-loss}

This appendix details how the differentiable JAX-Fluids solver~\cite{BezginBuhendwaAdams23,BezginBuhendwaAdams25} is used for physics-aware refinement of neural emulators.
We note that the solver, its discretization, the multi-block mesh, and the level-set-based immersed boundary method are described in \cref{subsec:jax-fluids}.

\paragraph{Residual loss.}
A steady-state solution of the compressible Euler equations, denoted by $\mathbf{U}^{*}$, satisfies
$$
  \mathbf{R} \left( \mathbf{U}^{*} \right) = \frac{\partial \mathbf{F}(\mathbf{U}^{*})}{\partial x}
  + \frac{\partial \mathbf{G}(\mathbf{U}^{*})}{\partial y} = 0
$$
where the convective flux vectors $\mathbf{F}$ and $\mathbf{G}$ are defined in \cref{app:phys_model} (see \cref{eq:convective_fluxes}).
Thus, at steady state, the divergence of the convective fluxes vanishes.
For a flow field $\mathbf{U}$ that does not satisfy the steady-state Euler equations exactly,
$$
  \mathbf{R} \left( \mathbf{U} \right) \neq 0.
$$

\paragraph{Point-wise residual calculation.}
In general, predictions of the physics emulator do not exactly satisfy the governing equations. We exploit this property to define a physics-based loss for fine-tuning. Given a surrogate prediction, we evaluate the discrete PDE residual using JAX-Fluids and minimize it by backpropagating through both JAX-Fluids and the physics emulator. 

The discrete residual vector in cell $(i,j)$ is given by
\begin{equation}
    \mathbf{R}_{i,j} \;=\;
    \frac{\mathbf{F}_{i+\frac{1}{2},j} - \mathbf{F}_{i-\frac{1}{2},j}}{\Delta x}
    \;+\;
    \frac{\mathbf{G}_{i,j+\frac{1}{2}} - \mathbf{G}_{i,j-\frac{1}{2}}}{\Delta y},
    \label{eq:pointwise-residual}
\end{equation}
where the cell-face fluxes $\mathbf{F}_{i\pm\frac{1}{2},j}$ and
$\mathbf{G}_{i,j\pm\frac{1}{2}}$ are computed with the same fifth-order WENO-Z
reconstruction and HLLC Riemann solver used during data generation, ensuring consistency of the residual operator with the discretization used for generating training data.
We note that cut cells contribute through their volume fraction and face apertures as outlined in \cref{app:phys_model}.

The point-wise residuals are aggregated into a scalar physics loss
$\mathcal{L}_{\mathrm{PDE}}$ by squaring each residual component, applying a per-equation weight $w_k$, and summing over all fluid cells,
\begin{equation}
    \mathcal{L}_{\mathrm{PDE}} = \sum_{(i,j) \in \Omega}
    \sum_{k=1}^{4} w_k \left( R^k_{i,j} \right)^{2} \Delta x \Delta y,
    \label{eq:phys-loss}
\end{equation}
where $R^{k}_{i,j}$ denotes the $k$-th component of $\mathbf{R}_{i,j}$,
corresponding to mass, $x$-momentum, $y$-momentum, and total energy. We found that the $y$-momentum had higher normalized residuals than the other conserved quantities and use
$w_k = (1,\,1,\,0.1,\,1)$, which down-weights its balance to ensure an equal representation of all components.

\paragraph{Differentiable coupling.}
Since the reference data were generated with JAX-Fluids, the refinement residual must be evaluated with the same solver: only then is the discrete residual consistent with the numerical discretization underlying the training data, so that minimizing it drives the prediction toward the solver solution rather than toward the fixed point of some other discretization. The surrogate, however, trains in PyTorch, so its predictions and the resulting gradients must be passed between PyTorch and the JAX-based solver. We bridge the two with zero-copy DLPack sharing wrapped in a custom autograd function, giving exact gradients consistent with the discretization. Floating-point inputs are upcast to double precision for the solver and the gradients are downcast back to the surrogate's working precision.

\paragraph{Refinement objective.}
Every refinement run starts from the same pre-trained ViT checkpoint and optimises a weighted sum of up to three terms,
$$
  \mathcal{L} = \mathcal{L}_\mathrm{data} + w_\mathrm{div}\,\mathcal{L}_\mathrm{div} + \lambda\,\mathcal{L}_\mathrm{PDE},
$$
namely a supervised data-reconstruction loss $\mathcal{L}_\mathrm{data}$ against the reference fields, a \emph{divergence} loss $\mathcal{L}_\mathrm{div}$ that penalises the mean-squared deviation of the prediction from a frozen copy of the pre-trained model, and the physics loss $\mathcal{L}_\mathrm{PDE}$.
$\mathcal{L}_\mathrm{PDE}$ and $\mathcal{L}_\mathrm{div}$ are target-free and refinement can therefore be applied to new samples without running a full simulation for each, only the mesh is needed.

A key advantage of the physics and divergence terms is that neither requires ground-truth fields. Both are evaluated from the model's own prediction and the mesh, so we can refine the surrogate on additional samples for which a simulation was never run and only a mesh and its design parameters are available. For these samples the supervised data term cannot be formed and is dropped.

The physics loss alone is not sufficient and leads to degenerate solutions. The divergence term prevents this by penalising deviation from the frozen pre-trained model, acting as a target-free replacement for the reconstruction loss, while the physics loss drives the field toward conservation consistency. This extends the surrogate to new regions of the design space at the cost of meshing rather than a full steady-state simulation.

Fine-tuning uses Lion (learning rate $10^{-6}$, weight decay $0.05$, gradient-norm clipping $1.0$) with a linear-warmup cosine-decay schedule in mixed (fp16) precision.
The variants we evaluate, which differ only in the active loss terms and the training data, are summarised in \cref{tab:refinement-variants}.

\begin{table}[h]
  \centering
  \caption{Refinement variants. All start from the same pre-trained ViT and share the optimiser and physics-loss settings above, differing only in the active loss terms and the training data ($D_1$ simulation data, $D_2$ simulation-free; see \cref{tab:datasets}). A dash denotes an inactive term. $\ast$ denotes fine-tuning on cases without existing field data, hence no data loss can be computed.}
  \vspace{.5em}
  \label{tab:refinement-variants}
  \begin{tabular}{lccccc}
    \toprule
    \textbf{Variant} & \textbf{Reconstruction} & \textbf{Divergence} & \textbf{Physics $\lambda$} & \textbf{Data} & \textbf{Epochs} \\
    \midrule
    Data only (baseline)         & $1$ & --    & --    & $D_1$            & $50$ \\
    Data + physics               & $1$ & --    & $0.2$ & $D_1$            & $50$ \\
    Divergence + physics             & --  & $0.2$ & $0.2$ & $D_1$            & $50$ \\
    $\text{Divergence + physics}^\ast$     & --  & $0.2$ & $0.2$ & $D_1\!+\!D_2$    & $25$ \\
    \bottomrule
  \end{tabular}
\end{table}

\section{Scaling Experiments}
\label{app:scaling}

\paragraph{Nested OOD splits with farthest-point sampling.} 
For data-scaling studies we additionally require training subsets of sizes ${n_1 < n_2 < \dots < n_K}$ that (i) share a single fixed (val, test, OOD) evaluation set and (ii) are strictly nested: $\mathcal{T}{n_1} \subset \mathcal{T}{n_2} \subset \dots \subset \mathcal{T}{n_K}$. 
After fixing the evaluation splits as above, we apply greedy farthest-point sampling (FPS) on the training pool. 
Parameters are min–max normalized to $[0,1]^d$ so that all dimensions contribute equally to Euclidean distance. 
FPS is initialized at the point closest to the pool centroid (to avoid boundary bias) and at every step appends $\arg\max_i \min{j \in \mathcal{S}} \lVert \mathbf{p}_i - \mathbf{p}_j \rVert_2$ to the selected set $\mathcal{S}$. 
Because each iteration grows $\mathcal{S}$ by exactly one point, every prefix of length $n_k$ is a valid $k$-center cover of the training pool, yielding the required nesting property. 
The training subset of size $n_k$ used in our scaling experiments is the first $n_k$ points of this FPS ordering. Validation, test, and OOD sets are identical across all $n_k$, isolating the effect of training-set size from variation in the evaluation distribution.
Every split is verified to (i) contain no duplicate indices within any subset, (ii) have empty pairwise intersection across subsets, and (iii) cover the full set of discovered run indices.

\paragraph{Hyperparameter searches.}
The scaling curves we report compare best-tuned configurations at every coordinate of the (model size, dataset size) grid rather than evaluating a single fixed recipe, to avoid systematically favoring any one capacity. 
For both the deterministic AB-UPT and ViT surrogates we ran a dedicated learning-rate sweep at every (model size, dataset size) cell of the scaling grid, bracketing the value reported in the implementation-details section by one order of magnitude on either side, and selected the run minimizing the in-distribution validation loss for inclusion in the reported curves. 
For the flow-matching ViT we additionally ran a grid over integration steps $N_\text{int}\in \{1,2,3,5,7,9,10,20,50,70\}$, and report the one that performs best on the validation set.

\section{Predictive Uncertainty}
\label{app:uncertainty}

\begin{figure}
    \centering

    \begin{subfigure}[t]{0.32\linewidth}\centering
        \includegraphics[width=\linewidth]{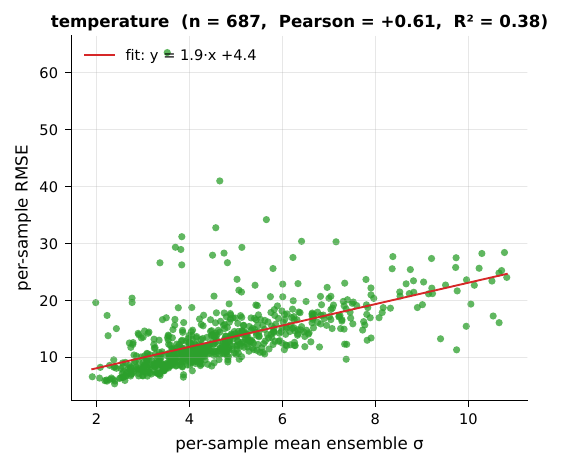}
        \caption{Temperature, flow matching}
    \end{subfigure}\hfill
    \begin{subfigure}[t]{0.32\linewidth}\centering
        \includegraphics[width=\linewidth]{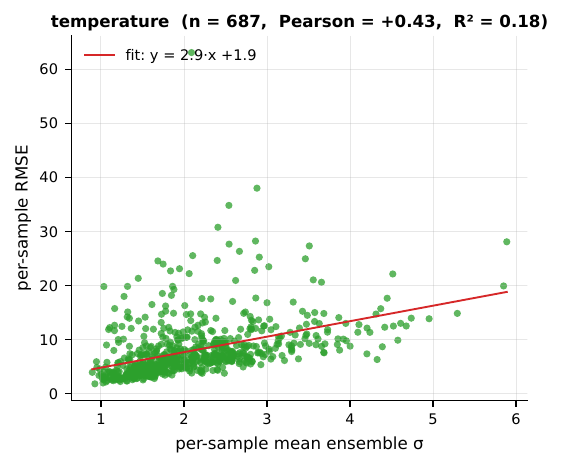}
        \caption{Temperature, ViT}
    \end{subfigure}\hfill
    \begin{subfigure}[t]{0.32\linewidth}\centering
        \includegraphics[width=\linewidth]{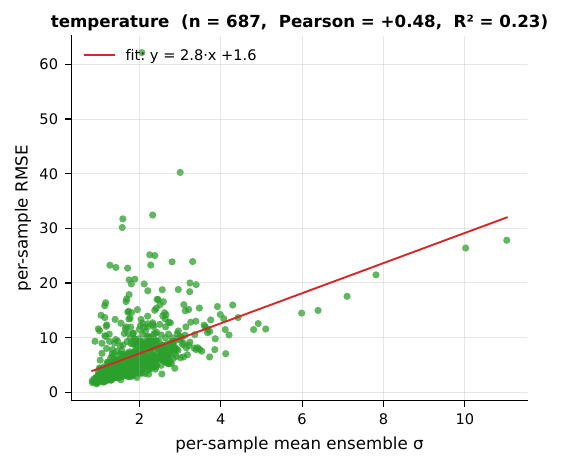}
        \caption{Temperature, AB-UPT}
    \end{subfigure}

    \vspace{0.4em}

    \begin{subfigure}[t]{0.32\linewidth}\centering
        \includegraphics[width=\linewidth]{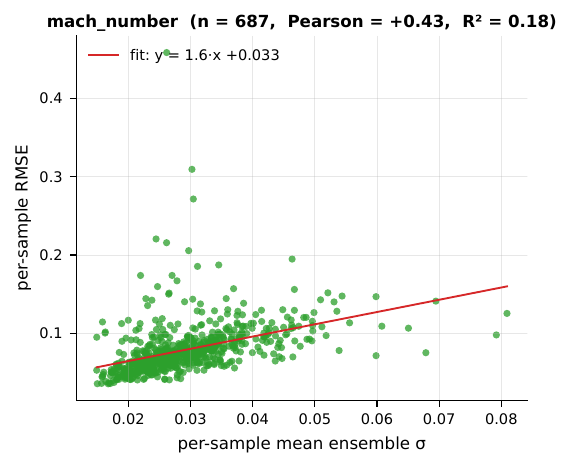}
        \caption{Mach number, flow matching}
    \end{subfigure}\hfill
    \begin{subfigure}[t]{0.32\linewidth}\centering
        \includegraphics[width=\linewidth]{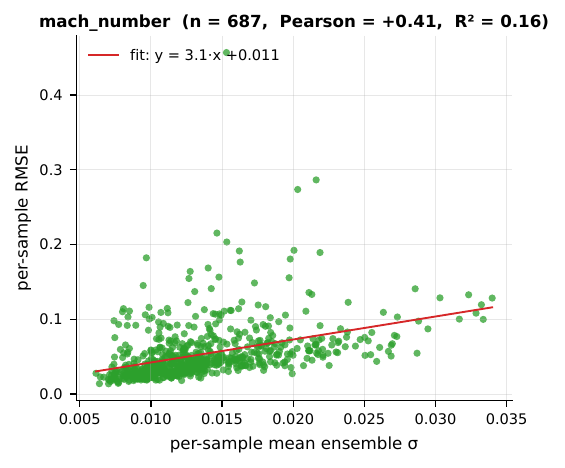}
        \caption{Mach number, ViT}
    \end{subfigure}\hfill
    \begin{subfigure}[t]{0.32\linewidth}\centering
        \includegraphics[width=\linewidth]{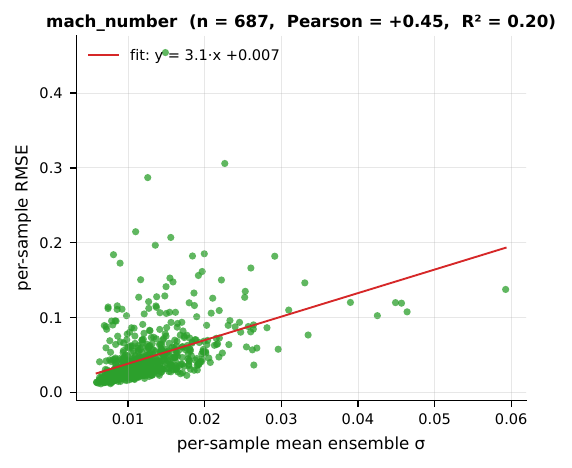}
        \caption{Mach number, AB-UPT}
    \end{subfigure}

    \vspace{0.4em}

    \begin{subfigure}[t]{0.32\linewidth}\centering
        \includegraphics[width=\linewidth]{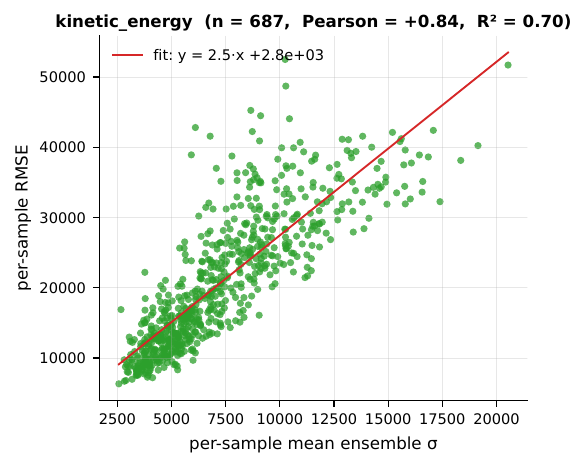}
        \caption{Kinetic energy, flow matching}
    \end{subfigure}\hfill
    \begin{subfigure}[t]{0.32\linewidth}\centering
        \includegraphics[width=\linewidth]{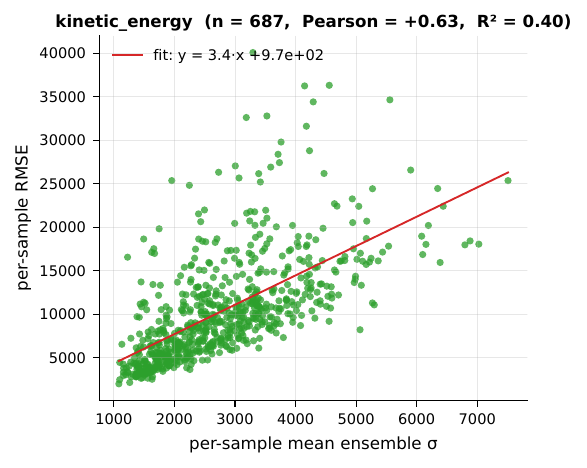}
        \caption{Kinetic energy, ViT}
    \end{subfigure}\hfill
    \begin{subfigure}[t]{0.32\linewidth}\centering
        \includegraphics[width=\linewidth]{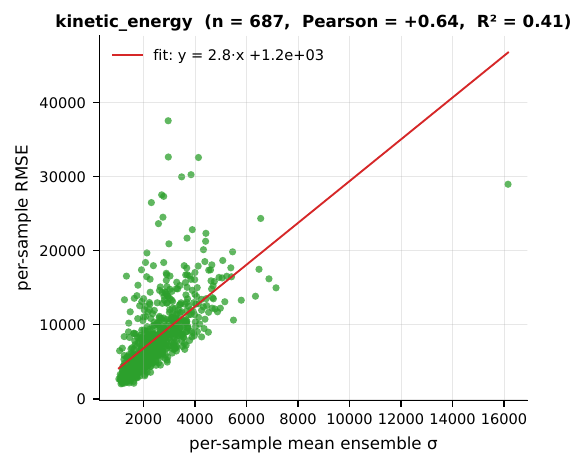}
        \caption{Kinetic energy, AB-UPT}
    \end{subfigure}

    \vspace{0.4em}

    \begin{subfigure}[t]{0.32\linewidth}\centering
        \includegraphics[width=\linewidth]{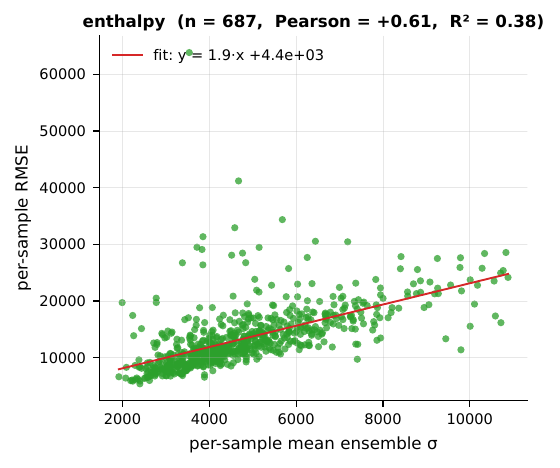}
        \caption{Enthalpy, flow matching}
    \end{subfigure}\hfill
    \begin{subfigure}[t]{0.32\linewidth}\centering
        \includegraphics[width=\linewidth]{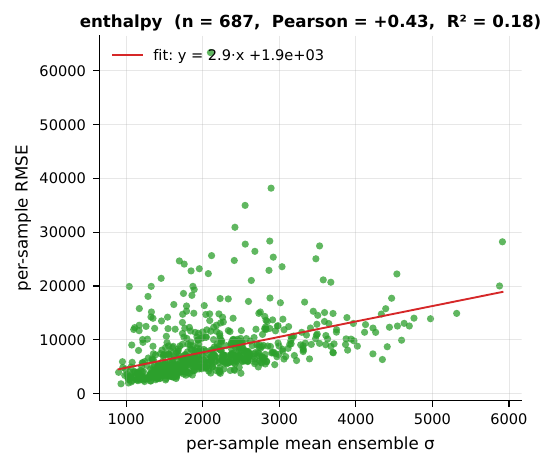}
        \caption{Enthalpy, ViT}
    \end{subfigure}\hfill
    \begin{subfigure}[t]{0.32\linewidth}\centering
        \includegraphics[width=\linewidth]{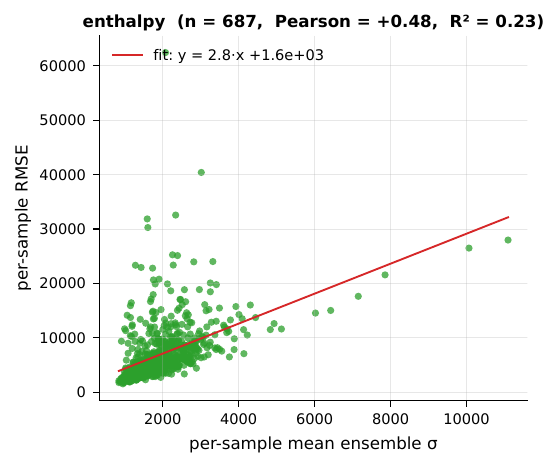}
        \caption{Enthalpy, AB-UPT}
    \end{subfigure}

    \caption{\textbf{Correlation of predictive uncertainty and error of emulator predictions for derived fields.}
    Rows correspond to %
    temperature (\textbf{a}--\textbf{c}),
    Mach number (\textbf{d}--\textbf{f}), kinetic energy (\textbf{g}--\textbf{i}) and enthalpy
    (\textbf{j}--\textbf{l}).
    Columns correspond, left to right, to flow matching, ViT and AB-UPT.
    All five fields are nonlinear algebraic combinations of the primitive variables and exhibit a correlation ranking that is consistent across the three surrogates with flow matching exhibiting best correlation between predictive error and uncertainty. 
    }
    \label{fig:calibration_derived_fields}
\end{figure}

\begin{figure}
    \centering
    \begin{subfigure}[t]{0.32\linewidth}\centering
        \includegraphics[width=\linewidth]{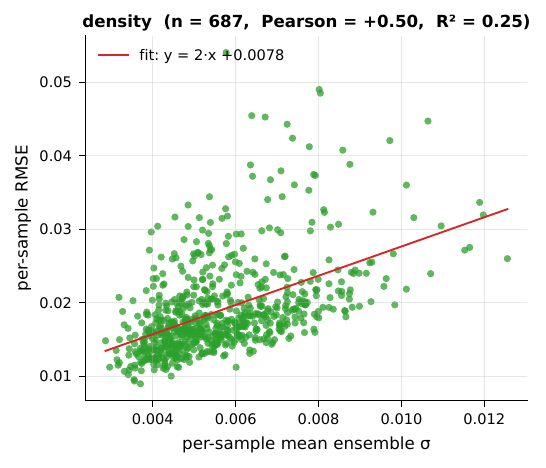}
        \caption{Density, flow matching}
    \end{subfigure}\hfill
    \begin{subfigure}[t]{0.32\linewidth}\centering
        \includegraphics[width=\linewidth]{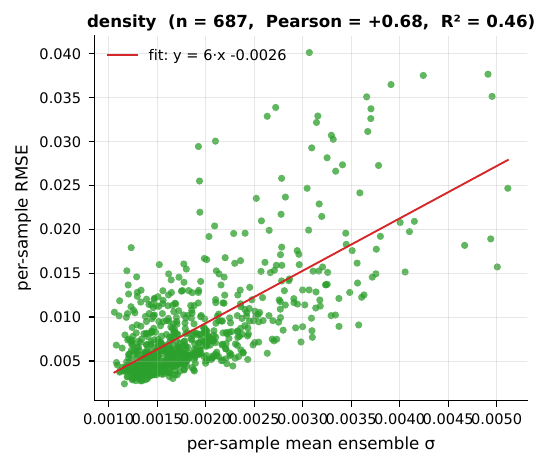}
        \caption{Density, ViT}
    \end{subfigure}\hfill
    \begin{subfigure}[t]{0.32\linewidth}\centering
        \includegraphics[width=\linewidth]{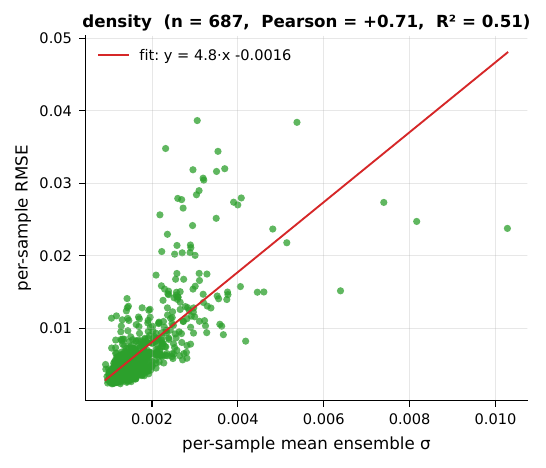}
        \caption{Density, AB-UPT}
    \end{subfigure}

    \vspace{0.4em}

    \begin{subfigure}[t]{0.32\linewidth}\centering
        \includegraphics[width=\linewidth]{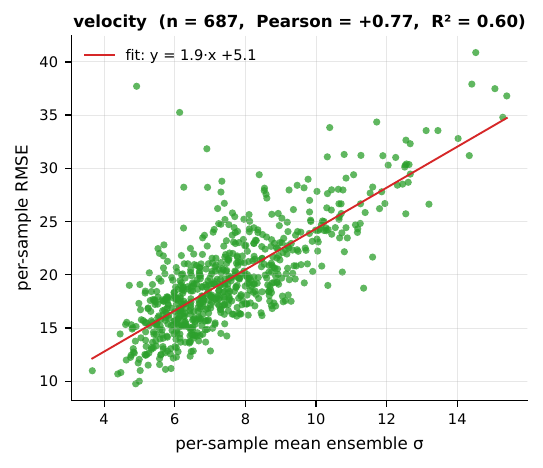}
        \caption{Streamwise Velocity,\\ flow matching}
    \end{subfigure}\hfill
    \begin{subfigure}[t]{0.32\linewidth}\centering
        \includegraphics[width=\linewidth]{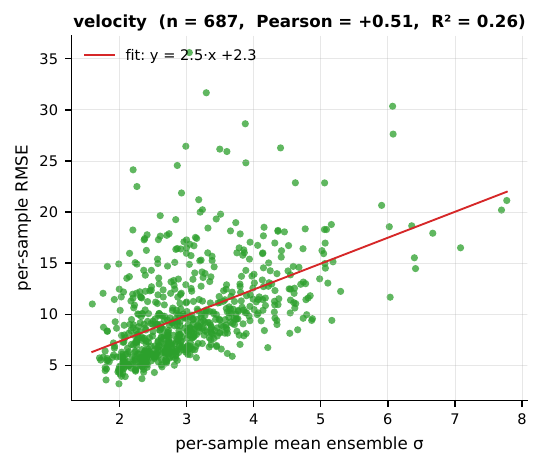}
        \caption{Streamwise Velocity,\\ ViT}
    \end{subfigure}\hfill
    \begin{subfigure}[t]{0.32\linewidth}\centering
        \includegraphics[width=\linewidth]{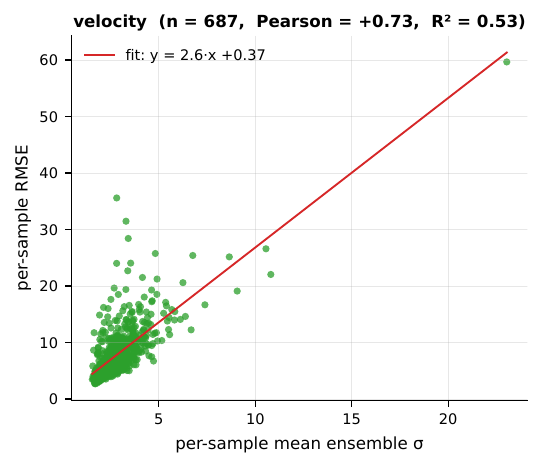}
        \caption{Streamwise Velocity,\\ AB-UPT}
    \end{subfigure}
    \caption{\textbf{Correlation of predictive uncertainty and error for density and velocity predictions for the three emulators.}
    Rows correspond to density (\textbf{a}--\textbf{c}), and streamwise velocity (\textbf{d}--\textbf{f}), columns correspond, left to right, to flow-matching,  ViT and AB-UPT.
    AB-UPT exhibits the best correlation and coefficient of determination between per-sample RMSE and per-sample average standard deviation as its inductive bias allows capturing position-wise shock and boundary layer artifacts. 
    }
    \label{fig:calibration_remaining_fields}
\end{figure}

We provide additional results for the correlation between predictive uncertainty and error across the three different methods on the remaining field predictions.
In \cref{fig:calibration_derived_fields} we show correlation on the temperature, Mach number, kinetic energy, and enthalpy fields.
Flow matching consistently attains the highest correlation and coefficient of determination compared to AB-UPT and ViT.
On the remaining density and velocity fields (see \cref{fig:calibration_remaining_fields}), the picture slightly changes.
Specifically, on those quantities AB-UPT exhibits the highest correlation between predictive error and uncertainty.
We believe the reason for this is that density and velocity contain sharp localized structures such as shocks and wall-aligned viscuous gradients, whereas the remaining fields are nonlinear algebraic combinations of those primitives and therefore smoother. 
AB-UPT is less affected by these artifacts because it learns a position-wise mapping from coordinates to field values and can therefore disentangle predictions.
ViT on the other hand uses patching with an initial average pooling layer such that the average coordinate per field is mapped to field values.
This design choice is deliberate to save computational complexity as smaller patch sizes yield longer token sequences that amplify complexity of the quadratic self-attention.
Therefore, we believe that these results are artifacts from design choices and do not reflect a direct disadvantage of regular-grid-based methods.

\section{Thermodynamic Self-consistency of Derived Quantities}
\label{app:thermodynamic_consistency}

We train the physics emulators to predict a redundant set of flow primitives aside from pressure $p$, density $\rho$, and velocity $\mathbf{u}$, which are thermodynamically independent.
The remaining fields follow in closed form from these primitives through their respective relations. In particular, 
the static temperature $T = p/(\rho R)$, the specific static enthalpy $h = \tfrac{\gamma}{\gamma-1} p/\rho$, the Mach number $Ma = |\mathbf{v}|/\sqrt{\gamma p/\rho}$, the total pressure $p_t = p (1 + \tfrac{\gamma-1}{2}Ma^2)^{\gamma/(\gamma-1)}$ and the kinetic-energy density $\mathrm{KE} = \tfrac{1}{2}\rho|\mathbf{v}|^2$, with $\gamma = 1.4$ and $R = 287.05~\mathrm{J\, kg^{-1}\, K^{-1}}$. 

We evaluate all emulators on the different splits. 
For every case the model's full inference path yields the predicted primitives on all fluid cells, with the flow-matching estimate taken as the ensemble mean over 10 independent stochastic samples.
For each derived quantity we report two errors against the ground truth, both as the relative $L_2$ norm taken over the fluid cells, namely the error of model's \emph{predicted} field, and the one of the \emph{derived} field, obtained by applying the relation above to the model's own predicted $p$, $\rho$, and $\mathbf{u}$.
Each entry of \cref{tab:pred_vs_derived_fields} lists predicted$,/,$derived (\%).
The agreement between both measures the thermodynamic self-consistency of an emulators predictions, while their difference quantifies whether emitting a redundant channel directly is preferable to deriving it from primitives.

Three patterns emerge. 
First, the temperature and enthalpy columns are identical, because $h$ and $T$ differ only by the constant factor $c_p = \gamma R/(\gamma-1)$.
As relative $L_2$ error is scale invariant the errors for both are consequently equal.
Second, direct prediction is consistently more accurate, oftentimes substantially, as it sidesteps error accumulation through a nonlinear transform.
The gap is mild for temperature, enthalpy and Mach number and negligible for the kinetic energy, indicating that the emulators' base predictions are very nearly thermodynamically self-consistent.
However, the gap is dramatic for the total pressure. 
Since $P_t$ depends on $\operatorname{Ma}$ through the steep isentropic power law, at the supersonic Mach numbers small velocity and Mach errors are amplified super-exponentially, inflating the derived total-pressure error to ${\sim}10^{4}\%$ for the ViT and ${\sim}10^{10}\%$ for flow matching, while the corresponding direct heads remain well behaved at $2.5$--$3.9\%$. 
Finally, the ranking across architectures is preserved on every split and all three degrade only modestly on the OOD split without changing order. 
Together these results show that the different emulators are thermodynamically coherent and argue for directly predicting derived quantities whose reconstruction from the base primitives is nonlinear, most critically the total pressure, but also the Mach number.

\begin{table}[tb!]
  \centering
  \caption{
  \textbf{Prediction vs deriving different physical quantities after pre-training.}
  Relative L2 errors (\%) for derived quantities enthalpy $h$, total pressure $p_t$, kinetic energy $k$, temperature $T$, and Mach number $\operatorname{Ma}$ for different models for predicting / deriving the different quantities.
  Predicting derived quantities directly generally results in smaller error.
  }
  \vspace{.5em}
  \label{tab:pred_vs_derived_fields}
  \setlength{\tabcolsep}{6pt}

  \resizebox{\textwidth}{!}{
  \begin{tabular}{llccccc}
  \toprule
  \textbf{Split} & \textbf{Model} &
   $h$ & $P_t$ & $k$ & $T$ & $\operatorname{Ma}$ \\
  \midrule

  \multirow{3}{*}{Val}
  & AB-UPT  & $1.56 / 1.74$ & $2.57 / 9.64$ & $1.99 / 1.99$ & $1.56 / 1.74$ & $0.90 / 1.14$ \\
  & ViT & $1.67 / 6.01$ & $2.66 / 6.52 \times 10^{4}$ & $2.17 / 2.17$ & $1.67 / 6.01$ & $0.97 / 2.54$ \\
  & Flow matching & $2.87 / 4.08$ & $3.44 / 3.67 \times 10^{10}$ & $4.56 / 4.58$ & $2.87 / 4.08$ & $1.42 / 6.50$ \\
  \midrule
  \multirow{3}{*}{Test}
  & AB-UPT & $1.53 / 1.72$ & $2.49 / 14.20$ & $1.96 / 1.95$ & $1.53 / 1.72$ & $0.88 / 1.12$ \\
  & ViT & $1.69 / 3.61$ & $2.61 / 5.83 \times 10^{4}$ & $2.18 / 2.19$ & $1.69 / 3.61$ & $0.96 / 2.48$ \\
  & Flow matching & $2.85 / 4.09$ & $3.45 / 6.2 \times 10^{10}$ & $4.50 / 4.51$ & $2.85 / 4.09$ & $1.42 / 7.19$ \\
  \midrule
  \multirow{3}{*}{OOD}
  & AB-UPT & $1.90 / 2.20$ & $2.93 / 15.49$ & $2.48 / 2.48$ & $1.90 / 2.20$ & $1.06 / 1.44$ \\
  & ViT & $2.06 / 12.96$ & $2.98 / 7.88 \times 10^{4}$ & $2.73 / 2.73$ & $2.06 / 12.96$ & $1.14 / 2.86$ \\
  & Flow matching & $3.25 / 4.80$ & $3.86 / 9.01 \times 10^{10}$ & $5.08 / 5.08$ & $3.25 / 4.80$ & $1.60 / 9.38$ \\
  \bottomrule

  \end{tabular}}
\end{table}

Finally, we also report the difference between predicting and deriving the different quantities after physics-aware model refinement in \cref{tab:pred_vs_derived_after_ref}.
Here we compare ViT before and after physics-aware model refinement.
Physics-aware refinement results in marginal improvements if the different field quantities are predicted directly.
However there is a significant reduction in error after refinement if the different quantities are derived.
This difference is particularly pronounced in derived fields that exhibit a nonlinear relation to primitive, like total pressure and Mach number.
For the former we sometimes even observe an improvement of around an order of magnitude.
This finding indicates that physics-aware model refinement facilitates thermodynamic consistency of the PE.

\begin{table}[tb!]
  \centering
  \caption{
  \textbf{Prediction vs deriving different physical quantities after physics-aware refinement.}
  Relative L2 errors (\%) for derived quantities enthalpy $h$, total pressure $p_t$, kinetic energy $k$, temperature $T$, and Mach number $\operatorname{Ma}$ for ViT before and after model refinement for predicting / deriving the different quantities.
  }
  \vspace{.5em}
  \label{tab:pred_vs_derived_after_ref}
  \setlength{\tabcolsep}{6pt}

  \begin{tabular}{llccc}
  \toprule
  & \textbf{Quantity} & \textbf{Val} & \textbf{Test} & \textbf{OOD} \\
  \midrule
  \multirow{5}{*}{Pre-training}
  & Enthalpy $h$ & $1.67 / 6.01$ & $1.69 / 3.61$ & $2.06 / 12.96$ \\
  & Total Pressure $p_t$ & $2.66 / 6.52 \times 10^{4}$ & $2.61 / 5.83 \times 10^{4}$ & $2.98 / 7.88 \times 10^{4}$ \\
  & Kinetic Energy $k$ & $2.17 / 2.17$ & $2.18 / 2.19$ & $2.73 / 2.73$ \\
  & Temperature $T$ & $1.67 / 6.01$ & $1.69 / 3.61$ & $2.06 / 12.96$ \\
  & Mach Number $\operatorname{Ma}$ & $0.97 / 2.54$ & $0.96 / 2.48$ & $1.14 / 2.86$ \\
  \midrule
  \multirow{5}{*}{Refinement}
  & Enthalpy $h$ & $1.66 / 3.66$ & $1.68 / 9.04$ & $2.03 / 4.64$\\
  & Total Pressure $p_t$ & $2.68 / 1.1 \times 10^{4}$ & $2.64 / 8.9 \times 10^{3}$ & $2.99 / 1.2 \times 10^{4}$ \\
  & Kinetic Energy $k$ & $2.16 / 2.36$ & $2.16 / 2.36$ & $2.69 / 2.82$ \\
  & Temperature $T$ & $1.66 / 3.66$ & $1.68 / 9.04$ & $2.03 / 4.64$ \\
  & Mach Number $\operatorname{Ma}$ & $0.96 / 1.55$ & $0.96 / 1.55$ & $1.12 / 1.73$\\
  \bottomrule
  \end{tabular}
\end{table}

\end{document}